\documentclass{article}

\usepackage{times}
\usepackage{graphicx} % more modern
\usepackage{subfigure} 
\usepackage[utf8]{inputenc}

\usepackage{amsmath,amsfonts}
\usepackage{dsfont}
\usepackage{stmaryrd}
\usepackage{cancel}
\usepackage{color}
\usepackage{makecell}

\usepackage{natbib}
\usepackage{algorithm}
\usepackage{algorithmic}

\usepackage{hyperref}

\usepackage{stmaryrd}
\usepackage{amsmath,amsfonts}
\usepackage{amssymb}
\usepackage{amsthm}
\usepackage{dsfont}

\DeclareMathOperator*{\argmax}{arg\,max}

\DeclareMathOperator*{\demi}{\frac{1}{2}}

\usepackage{comment}
\specialcomment{TUcom}{\color{red} $\langle\texttt{\bf TU}\rangle$ }{ $\langle\texttt{\bf \slash{}TU}\rangle$\color{black}}
\specialcomment{PGcom}{\color{blue} $\langle\texttt{\bf PG}\rangle$ }{ $\langle\texttt{\bf \slash{}PG}\rangle$\color{black}}
\usepackage{etoolbox}
\newtoggle{long-version}
\toggletrue{long-version}
\usepackage[accepted]{icml2015}

\icmltitlerunning{Relative Exponential Weighing for Dueling Bandits\iftoggle{long-version}{ (extended version)}{}}

\newtheorem{lemma}{Lemma}
\newtheorem{theorem}{Theorem}
\newtheorem{corollary}{Corollary}

\newlength\defaultparindent
\begin{document} 

\twocolumn[
\icmltitle{A Relative Exponential Weighing Algorithm for Adversarial Utility-based Dueling Bandits\iftoggle{long-version}{ (extended version)}{}}

% It is OKAY to include author information, even for blind
% submissions: the style file will automatically remove it for you
% unless you've provided the [accepted] option to the icml2014
% package.
\icmlauthor{Pratik Gajane}{pratik.gajane@orange.com}
\icmlauthor{Tanguy Urvoy}{tanguy.urvoy@orange.com}
\icmlauthor{Fabrice Clérot}{fabrice.clerot@orange.com}
\icmladdress{Orange-labs, Lannion, France}

% You may provide any keywords that you 
% find helpful for describing your paper; these are used to populate 
% the "keywords" metadata in the PDF but will not be shown in the document
\icmlkeywords{machine learning, ICML, bandits, learning from preference, online learning}

\vskip 0.3in
]

\begin{abstract} 
We study the \textit{K-armed dueling bandit problem} which is a variation of the classical Multi-Armed Bandit ({\sc mab}) problem in which the learner receives only relative feedback about the selected pairs of arms.
We propose an efficient algorithm called 
\textit{Relative Exponential-weight algorithm for Exploration and Exploitation}
({\sc rex3}) to handle the adversarial utility-based formulation of this problem.
We prove a finite time expected regret upper bound of order $\mathcal{O}(\sqrt{K\ln(K)T})$ for this algorithm and a general lower bound of order $\Omega(\sqrt{KT})$.
At the end, we provide experimental results using real data from information retrieval applications. %On other problems, our algorithm is competitive with the state of the art. % (PG)Made changes here.
\end{abstract} 

%%
%% TODO (PG)
%% check terms like: learner, player, algorithm, bandit, mab, dueling bandit,
%% and see if their usage is ambiguous or not.
%%

%% (PG)Checked and added a footnote. Only learner and algorithm mean the same in this context. bandit frequetly refers to a single instance of the problem.

\section{Introduction}

The \textit{K-armed dueling bandit problem} is a variation of the classical Multi-Armed Bandit ({\sc mab})
problem introduced by \citet{Yue:2009:IOI:1553374.1553527} 
to formalize the exploration/exploitation dilemma in learning from preference feedback.
In its utility-based formulation, at each time period, 
the environment sets a bounded value for each of $K$ arms. Simultaneously the learner selects two arms and wagers that one of the arms will be \textit{better} than the other.
The learner only sees the outcome of the \textit{duel} between the selected arms (i.e. the feedback indicates which of the selected arms has better value) and receives
the average of the rewards of the selected arms. The goal of the learner is to maximize her \textit{cumulative reward}.
The difficulty of this problem stems from the fact that the learning algorithm 
% FC: better replace learner by algorithm everywhere
%\footnote{The terms \textit{learner} and \textit{algorithm} have been used interchangeably} 
has no way of directly observing the reward of its actions. This is a perfect example of \textit{partial monitoring problem} as defined in \citet{conf/colt/PiccolboniS01}.
%and \citet{DBLP:journals/mor/BartokFPRS14}.

%In contrast to the classic {\sc mab} problem where the absolute reward of the chosen arm is returned as feedback, in the dueling bandits setting the feedback is the relative (pairwise) feedback of the two chosen arms (most of the time it is binary but it may be any non-decreasing function of the value difference).
Relative feedback is naturally suited to many practical applications like user-perceived product preferences, where a relative perception: \textit{``A is better than B''} is easier to obtain than its absolute counterpart: \textit{``A value is 42, B is worth 33''}.
Another important application of dueling bandits comes from information retrieval systems where users provide {\it implicit feedback} about the provided results. This implicit feedback is collected in various ways e.g. a click on a link, a tap, or any monitored action of the user. In all these ways however, this kind of feedback is often strongly biased by the model itself (the user cannot click on a link which was not proposed).
%%
%%TU: Any suggestion to express the same idea more formally? 
% made some changes above so the text seems more formal. 

To remove this bias in search engines, \citet{DBLP:conf/kdd/RadlinskiJ07} propose to interleave the outputs of different ranking models: the model which scores a click wins the duel. The accuracy of this \textit{interleave filtering method} was highlighted in several experimental studies \cite{DBLP:conf/kdd/RadlinskiJ07,Joachims:2007:EAI:1229179.1229181,Chapelle2012}.

% changes to present tense to maintain consistency. 

\subsection*{Main contribution}
%PG: Maybe we should have a separate subsection as "Main contribution" otherwise it's an abrupt jump from the previous paragraph to this paragraph. 
The main contribution of this article is an algorithm designed for the adversarial utility-based dueling bandit problem in contrast to most of the existing algorithms which assume a stochastic environment.

Our algorithm, called \textit{Relative Exponential-weight algorithm for Exploration and Exploitation} ({\sc rex3}), is a non-trivial extension of the 
\textit{Exponential-weight algorithm for Exploration and Exploitation}
({\sc exp3}) algorithm \cite{auer2002nonstochastic} to the dueling bandit problem.
We prove a finite time expected regret upper bound of order $\mathcal{O}(\sqrt{K\ln(K)T})$ and develop an argument initially proposed by \citet{DBLP:conf/icml/AilonKJ14} to exhibit a general lower bound of order $\Omega(\sqrt{KT})$ for this problem.

These two bounds correspond to the original bounds of the classical {\sc exp3} algorithm and the upper bound strictly improves from the $\tilde{\mathcal{O}}(K\sqrt{T})$ obtained by existing generic partial monitoring algorithms.\footnote{The notation $\tilde{\mathcal{O}}(\cdot{})$ hides logarithmic factors.}

Our experiments on information retrieval datasets show that the anytime version of {\sc rex3} is a highly competitive algorithm for dueling bandits, especially in the initial phases of the runs where it clearly outperforms the state of the art.
% FC: not the right place (i.e. when $t\leq 10^6$ for the most unfavorable dataset)
% Changed quotes to italic for algorithm name for consistency. At other places, we have used italic. 

\subsection*{Outline}

This article is organized as follows:
In section~\ref{sec:previous1}, we give a brief survey on dueling bandits with section~\ref{sec:adversarial} dedicated to the adversarial case.
Most notations and formalizations are introduced in this section.
Secondly, in section~\ref{sec:rex3}, we formally describe {\sc rex3} with its pseudo-code.
Then, in section~\ref{sec:upperbound}, we provide the upper bound on the expected regret of {\sc rex3}.
Furthermore, in section~\ref{sec:lowerbound}, we provide the lower bound on the regret of any algorithm attempting to solve the adversarial utility-based dueling bandit problem.
% and show that its is asymptotically equal to the corresponding lower bound for the classical {\sc mab} problem.
% PG: Why is the above statement is in the comment and not in the text?
Section~\ref{sec:xp} begins with an empirical study of the bound given in section~\ref{sec:upperbound}. It then provides comparisons of {\sc rex3} with state-of-the art algorithms on information retrieval datasets.
%FC: not the right place
% like {\sc rucb} \cite{zoghi2013relative}, {\sc savage} \cite{urvoy2013generic}, {\sc btm} \cite{Yue2011}, and  {\sc Sparring} when coupled with {\sc exp3} \cite{DBLP:conf/icml/AilonKJ14}.
The conclusion is provided in section~\ref{sec:conclusion}.
%and a few avenues for future research.

%\subsection{Previous work on multiarmed bandits}
\section{Previous Work and Notations}
\label{sec:previous1}

The conventional {\sc mab} problem has been well studied in the stochastic setting as well as the (oblivious) adversarial setting
\citep[see][]{Cesa-Bianchi:2006:PLG:1137817,DBLP:journals/ftml/BubeckC12}. 
%Some recent algorithms like
%\cite{conf/ewrl/SeldinSAA12} were also designed to cope with both stochastic and adversarial settings.
%\cite{journals/jmlr/BubeckS12,conf/ewrl/SeldinSAA12}.
%
These {\sc mab} algorithms are designed to optimize exploration and exploitation in order to control the \textit{cumulative regret} which is the difference between the gain of a reference strategy and the actual gain of the algorithm.
%\begin{equation}
%\label{eq:regret}
%R_T=\sum_{t=1}^T r_{alg}(t)=\sum_{t=1}^T x_{ref}(t)-x_{alg}(t)
%\end{equation}

\subsection{Exponential-weight algorithm for Exploration and Exploitation}
\label{sec:previous1bis}
Of particular interest are the \textit{Exponential-weight algorithm for Exploration and Exploitation} ({\sc exp3}) and its variants presented by \citet{auer2002nonstochastic} for the adversarial bandit setting.
For a fixed horizon $T$ and $K$ arms, the {\sc exp3} algorithm provides an expected cumulative regret bound of order $\mathcal{O}(\sqrt{K\ln(K)T})$ against the best single-arm strategy.
%\footnote{The algorithm {\sc inf} by \citet{AudibertINF2010} achieves the optimal $\mathcal{O}\left(\sqrt{KT}\right)$ bound.}.
This algorithm is indeed \textit{adversarial} because it does not require a stochastic assumption on the rewards. It is although not \textit{anytime} because it requires the knowledge of the horizon $T$ to run properly.
A \textit{``doubling trick''} solution is proposed by \citet{auer2002nonstochastic} to preserve the regret bound when $T$ is unknown. It consists of running {\sc exp3} in a carefully designed sequence of increasing \textit{epochs}. Another elegant solution was later proposed by \citet{conf/ewrl/SeldinSAA12} for the same purpose.

\subsection{Previous work on stochastic dueling bandits}
\label{sec:previous2}
The dueling bandits problem is recent, although related to previous works on computing with noisy comparison \citep[see for instance][]{Karp:2007:NBS:1283383.1283478}.
This problem also falls under the framework of \textit{preference learning} \cite{Freund:2003:EBA:945365.964285,LETORLiu2009,pre2010}
which deals with learning of (predictive) preference models from observed (or extracted) preference information i.e. relative feedback which specifies which of the chosen alternatives is preferred.
Most of the articles hitherto published on dueling bandits consider the problem under a stochastic assumption.

\citet{Yue:2009:IOI:1553374.1553527} propose an algorithm called \textit{Dueling Bandit Gradient Descent} ({\sc dbgd}) to solve
a version of the dueling bandits problem where context information is provided. They approach  ({\it contextual}) dueling bandits as an on-line convex optimization problem.
%: the arms are assumed to lie in a metric space and the duels outcome are assumed to be the noisy observations of the utility 
%gradient.

\citet{Yue/etal/12a} propose an algorithm called \textit{Interleaved Flitering} ({\sc if}).
Their formulation is stochastic and {\it matrix-based}:
for each pair $(i,j)$ of arms, there is an unknown probability $P_{i,j}$ for $i$ to win against $j$.
This  \textit{preference matrix} $P$ of size $K\!\times\!K$ 
must satisfy the following symetry property:
\begin{equation}
\label{eq:symetry}\forall i,j\in\{1,\ldots{},K\},\quad{}P_{i,j}+P_{j,i}=1
\end{equation}
Hence on the diagonal: $P_{i,i}= \demi \quad  \forall i \in \{1, \dots, K\}$.
Let $i^*$ be the ``best arm'' {(as we will see later, this best arm coincides with the notion of {\it Condorcet winner})}. \citet{Yue/etal/12a} define the regret incurred at the time instant $t$ when arms $a$ and $b$ are pulled as:
\begin{equation}
\label{eq:condorcetregret}
r'_{a,b}=\frac{P_{i^*,a}+P_{i^*,b}-1}{2}\quad\in{}(0,\demi{})
\end{equation}
We will call this regret a \textit{Condorcet regret}.

For the {\sc if} algorithm to work, the $P$ matrix is expected to satisfy several strong assumptions: {\it strict linear ordering}, {\it stochastic transitivity},
and {\it stochastic triangular inequality}.
Under these three assumptions {\sc if} is guaranteed to suffer an expected cumulative regret of order $\mathcal{O}(K\log{T})$.
%These assumptions constrain somehow the preference matrix to be the result of a utility-based generative model \citep[see][for details]{Yue/etal/12a}.

% switch to present tense here?

\citet{Yue2011} introduce \textit{Beat The Mean} ({\sc btm}), an algorithm which proceeds by successive elimination of arms.
%builds on the Successive Elimination algorithm proposed by \citet{Even-Dar2006}.
% for the conventional PAC bandit setting.
This algorithm is less constrained than {\sc if} as it also applies to a relaxed setting where the preference matrix can slighlty violate the stochastic transitivity assumption. Its cumulative regret bound is of order $\mathcal{O}(\gamma^7K\log{T})$ where $\gamma$ is here a known parameter.
%PG: Is this $\gamma$ same as the exploration/exploitation parameter we use later?
%TU: No but we may keep the notation or add a prime' for it only apears in a clear context.

\citet{urvoy2013generic} propose a generic algorithm called {\sc savage} (for Sensitivity Analysis of VAriables for Generic Exploration) which does away with several assumptions made in the previous algorithms e.g. existence of inherent values of arms, existence of a linear order amongst arms.
In this general setting, the {\sc savage} algorithm obtains a regret bound of order $\mathcal{O}(K^2\log{T})$.
The key notions they introduce for dueling bandits are the 
\textit{Copeland}, \textit{Borda} and \textit{Condorcet} scores \cite{DBLP:journals/anor/CharonH10}.
The Borda score of an arm $i$ on a preference matrix $P$ is $\sum_{j=1}^K P_{i,j}$ and its Copeland score is $\sum_{j=1}^K \llbracket{}P_{i,j}>\demi\rrbracket\ $
{(We use $\llbracket{}\ldots\rrbracket$ to denote the indicator function)}. 
If an arm has a Copeland score of $K-1$, which means that it defeats all the other arms in the long run, it is called a \textit{Condorcet winner}. The existence of a Condorcet winner is the minimum assumption required for the \textit{Condorcet regret} as defined on equation \eqref{eq:condorcetregret} to be applicable.
There exists however some datasets like \citet{MSLR30K} where this Condorcet condition is not satisfied.
%As we will see in Figure~\ref{fig:mslr30k}, in some experimental datasets from , there is no %Condorcet winner and some arms may provide negative regrets according to this definition.
%It is however possible to define a robust \textit{Copeland regret} which applies for any preference matrix.
% (PG)Shouldn't the above statements be in the main text?

\citet{zoghi2013relative} extend the \textit{Upper Confidence Bound} ({\sc ucb}) algorithm %\citep[see][]{DBLP:journals/ftml/BubeckC12}
\cite{Auer:2002:FAM:599614.599677} 
and propose an algorithm called \textit{Relative Upper Confidence Bound} ({\sc rucb}) provided that the preference matrix admits a Condorcet winner.
They retrieve an $\mathcal{O}(K\log{T})$ bound under this sole assumption.
Unlike the previous algorithms, {\sc rucb} is an anytime dueling bandits algorithm since it does not require the time horizon $T$ as input.

%An interesting work by 
\citet{DBLP:conf/icml/AilonKJ14} propose three methods ({\sc Doubler}, {\sc Multismb}, and {\sc {\sc Sparring}}) to reduce the stochastic \textit{utility-based dueling bandits} problem to the conventional {\sc mab} problem.
A stochastic dueling bandits problem is \textit{utility-based} if the preference is the result of comparisons of the individual utility/reward of the arms.
This is a strong restriction from the general -- {\it matrix-based} -- formulation of the problem.

More formally, there are $K$ probability distributions $v_1,\dots,v_K$ over $[0,1]$
%I think this should be a closed interval [] as values could be 0 or 1. 
 associated respectively with arms $1,\dots,K$. Let $\mu_1, \dots, \mu_K$ be the respective means of $v_1, \dots, v_K$. When an arm $a$ is pulled, its reward/utility $x_a$ is drawn from the corresponding distribution  $v_a$\footnote{Note that we frequently drop the time index when it is unnecessary or clear from context. For instance we simply write $x_a(t)$ or simply $x_{a}$ instead of $x_{a_t}(t)$.
%The $i,j$ notations will always refer to static indices.
}.
%the (an, in case of more than one) TU: -> ICML reviewers know that fact
Let $i^*\in\argmax \mu_i$ be an optimal arm. The regret incurred at the time instant $t$ when arms $a$ and $b$ are pulled is defined as:
\begin{equation}
\label{eq:banditregret}
r_{a,b}(t) = \frac{2x_{i^*} - x_a - x_b}{2}
\end{equation}
We will call this regret a \textit{bandit regret}.
With randomized tie-breaking we can rebuild the preference matrix:
\begin{align*}
P_{i,j}&=\mathds{P}(x_i>x_j)+\demi{}\mathds{P}(x_i=x_j)
\end{align*}
When all $v_i$ are Bernoulli laws, this reduces to:
\begin{align}
\label{eq:tiebreak}P_{i,j}&=\frac{\mu_i - \mu_j + 1}{2}
\end{align}
Note that if $\mu_i>\mu_j$ for some arms $i$  and $j$, then $P_{i,j}>\demi{}$. The best arm in the usual bandit sense hence coincides with the Condorcet winner (which turns out to be the Borda winner too) on the matrix formulation and the expected bandit regret is twice the Condorcet regret as defined in \eqref{eq:condorcetregret}:
$$
\mathds{E}r_{a,b}=\frac{2\mu_{i^*}-\mu_{a}-\mu_{b}}{2}={P_{i^*,a}+P_{i^*,b}-1}=2r'_{a,b}
$$

Several other models and algorithms have been proposed since to handle stochastic dueling bandits. We can cite \cite{conf/icml/Busa-FeketeSCWH13,icml2014c2_busa-fekete14,zoghi2014relative,zoghi2015mergerucb}.
See also \cite{busa2014survey} for an extensive survey of this domain.

%\section{Adversarial dueling bandits}
\subsection{Adversarial dueling bandits}
\label{sec:adversarial}
The bibliography on stochastic dueling bandits is flourishing, but the results about adversarial dueling bandits remain quite scarce.

A utility-based formulation of the problem is however proposed in
\citet[section~6]{DBLP:conf/icml/AilonKJ14}.
In this setting, as in classical adversarial {\sc mab}, the environment chooses beforehand an horizon $T$ and a sequence of utility/reward vectors $\mathbf{x}(t)=(x_1(t),\ldots,x_K(t)) \in [0,1]^{K}$ for $t=1,\ldots,T$.
The learning algorithm aims at controling the bandit regret against the best single-arm strategy, as defined in \eqref{eq:banditregret}, by choosing properly the pairs of arms $(i,j)$ to be compared.

To tackle this problem, \citet{DBLP:conf/icml/AilonKJ14} suggest to apply the {\sc Sparring} reduction algorithm, although originally designed for stochastic settings, with an adversarial bandit algorithm like {\sc exp3} as a black-box {\sc mab}. According to the authors, the {\sc Sparring} reduction preserves the $\mathcal{O}(\sqrt{KT\ln{}K})$ upper bound of {\sc exp3}.
%Our experiments on section~\ref{sec:xp} confirms that it is a sound idea.
%On Section~\ref{sec:lowerbound} we indeed develop their reduction argument to exhibit an $\Omega(\sqrt{KT})$ lower bound for the adversarial utility-based dueling bandits problem.
%No demonstration of this statement is however given in the paper.
This algorithm uses two separate {\sc mab}s (one for each arm). As a consequence, when it gets a relative feedback about a duel $(i,j)$, the left instantiation of {\sc exp3} only updates its weight for arm $i$ while the right instantiation only updates for $j$. The algorithm we propose improves from this solution by centralizing information for both arms on a single weight vector.

{%\color{red}

%\subsection*{Utility-based dueling bandits as a partial monitoring game}
As mentionned earlier, the dueling bandits problem is a special instance of a partial monitoring game \cite{conf/colt/PiccolboniS01,Cesa-Bianchi+Lugosi:2009,DBLP:journals/mor/BartokFPRS14}.
%% Pratik 
%\subsubsection*{Dueling bandits as a partial monitoring problem}
A partial monitoring game is defined by two matrices: a loss matrix $\mathcal{L}$ and a feedback matrix $\mathcal{H}$. These two matrices are known by the learner.
At each round, the learner chooses an action $a$ while the environment simultaneously chooses an outcome (say $\mathbf{x}$).
The learner receives a feedback $\mathcal{H}(a,\mathbf{x})$ and suffers (in a blind manner) a loss $\mathcal{L}(a,\mathbf{x})$.

It is straightforward to encode the classical {\sc mab} as a finite partial monitoring game: the actions are arms indices $a\in\{1,\ldots,K\}$ while the environment outcomes are reward vectors $\mathbf{x}(t)=(x_1(t),\ldots,x_K(t)) \in [0,1]^{K}$.
The loss and feedback matrices are respectivly defined by 
$\mathcal{L}(a,\mathbf{x})=-x_a$ and 
$\mathcal{H}(a,\mathbf{x})=x_a$.
For the {utility-based dueling bandits} problem, the learner's actions are the duels $(a,b)\in \{1,\ldots,K\}^2$ and the environment outcomes are reward vectors. If we constrain the rewards to be binary it turns out to be a finite partial monitoring game.
The Loss matrix is defined by $\mathcal{L}\left((a,b),\mathbf{x}\right)={-(x_a + x_b)}/{2}$ and the feedback by
$\mathcal{H}\left((a,b),\mathbf{x}\right)=\psi( x_a - x_b)$ where $\psi$ is a non-decreasing transfer function such that $\psi(0)=0$ (usually $\psi(x)=\llbracket x>0 \rrbracket$ or $\psi(x)=x$).
\iftoggle{long-version}{

There are only four classes of finite partial monitoring games in term of time lower bounds, namely: {\it trivial games} with no regret at all, {\it ``easy'' games} with $\tilde{{\Theta}}\left(\sqrt{T}\right)$ minimax regret, {\it ``hard'' games} with $\tilde{{\Theta}}\left(T^{2/3}\right)$, and {\it hopeless' games} with ${{\Omega}}\left(T\right)$ regret.

}

Several generic partial monitoring algorithms were recently proposed for both stochastic and adversarial settings \citep[see][for details]{DBLP:journals/mor/BartokFPRS14}. If we except  {\sc globalexp3} \cite{bartokColt13} which tries to capture more finely the structure of the games, these algorithms only focus on the time bound and perform inefficiently when the number of actions grows.

In a dueling bandit the number of non-duplicate actions is actually $K(K+1)/2$ and these algorithms, including {\sc globalexp3}, provide at best a $\tilde{\mathcal{O}}\left(K\sqrt{T}\right)$ regret guarantee. The dedicated algorithm that we propose is using the preference feedback more efficiently.

%the expected cumulative regret/loss for a direct application of this algorithm on dueling bandits is $O(KT^{2/3}\sqrt{\ln{T}})$.}

%\citet{conf/colt/PiccolboniS01} propose a {generic algorithm}
%called {\sc FeedExp3} to solve finite partial monitoring games. The crucial assumption for {\sc FeedExp3} to work is that there exists a matrix  ${\mathcal{K}}$ such that $\mathcal{L}= {K}\cdot\mathcal{H}$. 
%{Please refer to \cite{conf/colt/PiccolboniS01} or \citet[Chapter~6]{Cesa-Bianchi:2006:PLG:1137817} for a detailed discussion about this algorithm and the partial monitoring problem. With the relevant parameters set accordingly, the expected cumulative regret/loss for a direct application of this algorithm on dueling bandits is $O(KT^{2/3}\sqrt{\ln{T}})$.}
%This bound is however clearly suboptimal.

%Recent works by \citet{DBLP:journals/mor/BartokFPRS14} and \cite{bartokColt13} 
%improve from {\sc FeedExp3} by introducting a hierarchy of finite problems with new generic algorithms, but the application of these algorithms to dueling bandits is not trivial.
%Even if we assume binary rewards, 
}

\section{Relative Exponential-weight Algorithm for Exploration and Exploitation}
\label{sec:rex3}
The pseudo-code for the algorithm we propose is given in Algorithm~\ref{alg:rex3}.
As previously stated on Section~\ref{sec:adversarial}, this algorithm is designed to apply for the adversarial utility-based dueling bandits problem.

It is similar to the original {\sc exp3} 
from step~1 to step~6 where it
computes a distribution $\mathbf{p}(t)=\left(p_1(t),\ldots,p_K(t)\right)$ 
which is a mixture of a normalized weighing of the arms $w_i/\sum_iw_i$ and a uniform distribution $1/K$. As in {\sc exp3}, this uniform probability is introduced to ensure a minimum exploration of all arms.

At step~7, the algorithm draws two arms $a$ and $b$ independently according to $\mathbf{p}(t)$.
At step~8, the algorithm gets $\psi(x_{a}-x_{b})$ as relative feedback .
Note that, since arms are drawn with replacement, we may have $a=b$, in which case the algorithm will get no information.
This event is indeed expected to become frequent when the $\mathbf{p}(t)$ distribution becomes peaked around the best arms.
This necessity for a regret-minimizing dueling bandits algorithm to renounce getting information when confident about its decision is a structural bias toward exploitation that is not encountered in classical bandits.

Step~8 is the big difference from {\sc exp3}; because we only have access to the relative $\psi(x_{a}-x_{b})$ value, we have no mean to estimate the individual rewards $x_a$ or $x_b$.
There is however a solution to circumvent this problem:
the best arm in expectation at time $t$ is not only the one which maximizes the absolute reward. It is indeed the one which maximizes the regret of any fixed strategy $\pi(t)$ against it: 
$$\argmax_i x_i(t)=\argmax_i{\left(x_i(t)-\mathds{E}_{a\sim{}\pi(t)}x_a\right)}\ .$$
This reference strategy could be a single-arm or uniform strategy but playing a suboptimal strategy to get a reference has a cost in terms of regret. One of our contributions is to show that the algorithm may use its own strategy as a reference.
  
At step~9, the condition $a\neq b$ is only a slight improvement for matrix-based dueling bandits where the outcome of a duel of an arm against itself is randomized as in \eqref{eq:tiebreak}.
%(i.e. when $\mathds{P}(a \text{ wins against } a)=\demi$).

At steps 10 and 11, the weights of the played arms 
are updated. This update process is the core of our algorithm, it will be detailed in Section~\ref{sec:analysis}.

Step~13 is only required for the anytime version of the algorithm. It will be explained in section~\ref{sec:xp2}.

\begin{algorithm}
\caption{{\sc rex3}: Exp3 with relative feedback}
\label{alg:rex3}
\begin{algorithmic}[1]
 \STATE \textbf{Parameters:} $\gamma \in (0,1]$ \\
 \STATE  \textbf{Initialization:} $w_i(1) = 1$ for $i=1,\dots, K$. \\
\FOR{$t=1,2,\dots$}  
\label{alg:step0}\FOR{$i=1,\dots,K$}  
\STATE Set $p_i(t) \leftarrow (1 - \gamma) \frac{w_i(t)}{\sum_{j=1}^{K}w_j(t)} + \frac{\gamma}{K}$
\label{alg:step1}\ENDFOR
%% TU: We do not need the `with replacement'
%% as `without replacement' cannot be independent.
\label{alg:step2}\STATE Pull two arms $a$ and $b$ chosen independently according to the distribution $\left(p_1(t), \dots, p_K(t)\right)$.
\label{alg:step3}\STATE Receive relative feedback $\psi(x_a-x_b)$ $\in [-1,+1]$
%\STATE For $i=1,2,\dots,K$ set \\
% \[\hat{c}_i(t) = \llbracket{}i = a\rrbracket{} \frac{x_a - x_b}{2p_a} + \llbracket{}i = b\rrbracket{} \frac{x_b - x_a}{2p_b}\]
% \[w_i(t+1) = w_i(t) e^{((\gamma/K) \hat{c}_i(t))}\]
\label{alg:step4}\IF{$a\neq b$}
\STATE Set $w_a(t+1) \leftarrow w_a(t) \cdot{} e^{\frac{\gamma}{K}\frac{\psi(x_a - x_b)}{2p_a}}$
\STATE Set $w_b(t+1) \leftarrow w_b(t) \cdot{} e^{-\frac{\gamma}{K}\frac{\psi(x_a - x_b)}{2p_b}}$
\label{alg:step5}\ENDIF
\STATE \textit{Update $\gamma$ (for anytime version)}
\ENDFOR
\end{algorithmic}
\end{algorithm}

\section{Analysis}
\label{sec:analysis}
For the analysis, we focus on the simple case where $\psi$ is the identity. It provides a
ternary {\it win/tie/loss} feedback if we assume binary rewards.
%This transfer function is simple and behaves simply in expectation.
%If we assume binary rewards it gives a ternary win/tie/loss feedback.
The main difference between {\sc exp3} and our algorithm is at steps 10 and 11 of Algorithm~\ref{alg:rex3}, where we update the weights according to the duel outcome: the winning arm is gratified while the loser is penalized. This `punitive' approach of exponential weighing departs from  {\sc exp3} and other weighing algorithms which gratify the most rewarding arms while kindly ignoring the non-rewarding ones \cite{freund1999adaptive,Cesa-Bianchi:2006:PLG:1137817}.

\subsection{Upper bound for {\sc rex3}}
\label{sec:upperbound}
In this section, we provide a finite-horizon non-stochastic upper bound on the expected regret against the best single action policy.

The steps 10-11 on Algorithm~\ref{alg:rex3} are equivalent to operating for each arm $i$ an update of the form:
$$w_i(t+1) = w_i(t)\cdot{}e^{\frac{\gamma}{K}\hat{c}_i(t)}$$
where
\begin{equation}
\label{eq:chat}
\hat{c}_i(t) = \llbracket{}i = a\rrbracket{} \frac{\psi(x_a - x_b)}{2p_a} + \llbracket{}i = b\rrbracket{} \frac{\psi(x_b - x_a)}{2p_b}
\end{equation}
One big difference with {\sc exp3} is that $\hat{c}_i(t)$ not an 
estimator of the reward $x_i(t)$. We instead have:
\begin{lemma}  
\label{lem:expectC}
$$ \mathds{E}\left[\hat{c}_i(t)|(a_1,b_1),..,(a_{t-1},b_{t-1})\right]
= \mathds{E}_{a\sim{}p(t)} \psi(x_i(t) - x_a(t))
$$
\end{lemma}
\iftoggle{long-version}{\textit{The proof of the lemma is given in Appendix~\ref{sec:expectC}.}
}{}

If $\psi$ is the identity then $\mathds{E} \hat{c}_i(t)
= x_i(t) - \mathds{E}_{a\sim{}p(t)}x_a(t)$ in which case we estimate the expected instantaneous regret of the algorithm against arm $i$.
If we rather take $\psi(x)=\llbracket{}x>0\rrbracket{}$, then
$\mathds{E}\hat{c}_i(t)
= \mathds{P}_{a\sim{}p(t)}\left(x_i(t)>x_a(t)\right)$, i.e. the probability for the algorithm to select an arm defeated by $i$.

Let $\mathbb{G}_{max}=\max_i\sum_{t=1}^{T}x_i(t)$ be the best single-arm gain, and let $\mathbb{G}_{alg}=\demi\sum_{t=1}^{T}x_a(t)+x_b(t)$
be the gain of the algorithm.
Let $\mathds{E}\mathbb{G}_{unif}=\frac{1}{K}\sum_{t=1}^{T} \sum_{i=1}^Kx_i(t)$ be the average value of the game (i.e. the expected gain of the uniform sampling strategy).

\begin{theorem}
\label{thm:mainTheorem}
If the transfer function $\psi$ is the identity and $\gamma\in (0,\demi)$, 
then,
$$\mathbb{G}_{max} - \mathds{E}(\mathbb{G}_{alg}) \leq \frac{K}{\gamma}\ln(K) + \gamma \tau$$
where
$\tau=
{e}\cdot\mathds{E}\mathbb{G}_{alg}
-\left(4\!-\!e\right)\cdot\mathds{E}\mathbb{G}_{unif}$.
%\begin{align}
%\tau&={e}\mathbb{G}_{max}
%-\left(4\!-\!e+(e\!-\!2)\varepsilon\right)\mathbb{G}_{min}
%\end{align}
%$\tau = (e - \frac{2\epsilon(e-2)}{K})G_{max} - \left( \frac{4-e}{2K} \right) G_{min}$.
\end{theorem}
\iftoggle{long-version}{\textit{The proof of this theorem is detailed on Appendix~\ref{sec:thmproof}.}}{We give a sketch of proof for this result in 
section~\ref{sec:thmproofsk}.}

Provided that $\mathds{E}\mathbb{G}_{alg}\leq{}\mathbb{G}_{max}$
 and $\mathds{E}\mathbb{G}_{unif}\geq \mathbb{G}_{min}$,
where $\mathbb{G}_{min}=\min_i\sum_{t=1}^{T}x_i(t)$ is the gain of the worst single-arm strategy, we can simplify the bound into:
\begin{corollary}
\label{thm:mainTheorem2}
\begin{align*}
&\mathbb{G}_{max} - \mathds{E}\mathbb{G}_{alg}
&\leq
\frac{K\!\ln{}K}{\gamma} 
+\gamma\left({e}\mathbb{G}_{max}
-\left(4\!-\!e\right)\mathbb{G}_{min}\right)
\end{align*}
\end{corollary}
As in \citep[section~3]{auer2002nonstochastic}, since $\frac{K}{\gamma}\ln(K) + \gamma \tau$ is  convex, we can obtain the optimal $\gamma$ on $(0,\demi)$:
\begin{align}
\label{eq:optigamma}\gamma* &= \min \left\{ \demi, \sqrt{\frac{K\ln(K)}{\tau}} \right\}
\end{align}
%On Figure~\ref{fig:lem2sens}, we provide an empirical study of Corollary~\ref{thm:mainTheorem2} bound and its relevance for the optimal $\gamma$ value determination (see discussion on Section~\ref{sec:xp1}).

%\subsection{Bound for {\sc rex3}}
Substituting $\gamma$ in Corollary~\ref{thm:mainTheorem2} with its optimal value from eq. \eqref{eq:optigamma} we obtain:
\begin{align*}
%\mathbb{G}_{max} - \mathds{E}(\mathbb{G}_{alg}) &\leq \frac{K}{\sqrt{\frac{K\ln(K)}{\tau}}}\ln(K) + \sqrt{\frac{K\ln(K)}{\tau}}\tau \\
%\mathbb{G}_{max} - \mathds{E}(\mathbb{G}_{alg}) &\leq \sqrt{K\ln(K)\tau} + \sqrt{K\ln(K)\tau} \\
\mathbb{G}_{max} - \mathds{E}(\mathbb{G}_{alg}) &\leq 2\sqrt{K\ln(K)\left[{e}\mathbb{G}_{max}
-\left(4\!-\!e\right)\mathbb{G}_{min}\right]}
\end{align*}
Hence:
\begin{corollary}
\label{thm:mainTheorem3}
When $\gamma = \min \left\{\demi, \sqrt{\frac{K\ln(K)}{\tau}} \right\}$, the expected regret of {\sc rex3} (Algorithm~\ref{alg:rex3}) is $\mathcal{O}\left(\sqrt{K\ln(K)T}\right)$.
\end{corollary}
The upper bound of {\sc rex3} for adversarial utility-based dueling bandits is hence the same as the one of {\sc exp3} for aversarial {\sc mab}s.
For a high-number of arms or a short term horizon, this bound is competitive against
the $\mathcal{O}\left(K\ln(T)\right)$ or $\mathcal{O}\left(K^2\ln(T)\right)$ existing bounds for stochastic dueling bandits.

\subsection{Lower bound for dueling bandits algorithms}
\label{sec:lowerbound}
To provide a lower bound on the regret of any dueling bandits algorithm, we use a reduction
to the classical {\sc mab} problem suggested by \citet{DBLP:conf/icml/AilonKJ14}.
\begin{algorithm}
\caption{Reduction to classical {\sc mab}}
\label{alg:exp3simu}
\begin{algorithmic}[1]
\STATE DBA.{\tt{}init()}
\STATE Set $t=1$
\REPEAT
\STATE $(a_t, b_{t+1}) \leftarrow$ DBA.{\tt{}decide()}
\STATE $x_{a_t}  \leftarrow$ CBE.{\tt{}get\_reward()}
\STATE $x_{b_{t+1}} \leftarrow$ CBE.{\tt{}get\_reward()}
\STATE DBA.{\tt{}update($(a_t, b_{t+1}), (x_{a_t}- x_{b_{t+1}}) $)} 
\STATE $t = t + 2$ 
\UNTIL{$t\geq{}T$}
\end{algorithmic}
\end{algorithm}

Algorithm~\ref{alg:exp3simu} gives an explicit formulation of this reduction
by using a generic dueling bandits algorithm (DBA) as a black-box having the following public sub-routines: {\tt init()}, {\tt decide()} and {\tt update()}.
%The subroutine {\tt init()} is used to initialize the algorithm, {\tt decide()} returns the pair of arms to be pulled at any given time instant and {\tt update()} receives the relative feedback.
The classical bandit environment (CBE) provides {\tt get\_reward()} which returns the reward of the input arm.
%% TU: to be kept for long version
%By definition, the gain obtained in the dueling bandits setting on selection of the pair $(a_t, b_{t+1})$ is $\frac{x_{a_t} + x_{b_t+1}}{2}$.
%On the other hand, according to the definition of gain in the classical bandit setting, the gain obtained on selection of arm $a_t$ followed by arm $b_{t+1}$ is $x_{a_t} + x_{b_t+1}$.
%So clearly the expected classical-bandit gain of Algorithm~\ref{alg:exp3simu} will be twice to expected gain of the black-box dueling bandits it uses.
The expected classical-bandit gain of Algorithm~\ref{alg:exp3simu} will be twice the expected gain of the black-box dueling bandits it uses.

It is important to note that this reduction only works for stochastic settings where the expected reward of each arm remains the same across time instants.
%because the rewards are drawn from stationary distributions.
According to Theorem $5.1$ given by \citet[section~5]{auer2002nonstochastic}, for $K\geq2$, the expected regret in the classical bandit setting is $\Omega(\sqrt{KT})$ (assuming $T$ is large enough i.e. $T \geq \sqrt{KT}$). 
Since this result is obtained with a stationary stochastic distribution, by extension, the expected regret in the dueling bandits setting cannot be less than $\Omega(\sqrt{KT})$.

\begin{theorem}
\label{thm:lowerbound}
For any number of actions $K\geq 2$ and large enough time horizon $T$ (i.e. $T \geq \sqrt{KT}$), there exists a distribution over assignments of rewards such that the expected cumulative regret of any utility-based dueling bandits algorithm cannot be less than $\Omega(\sqrt{KT})$.
\end{theorem}

\section{Experiments}
\label{sec:xp}
To evaluate {\sc rex3} and other dueling bandits algorithms, we 
have applied them to the online comparison of rankers for search engines by \textit{interleaved filtering} \cite{DBLP:conf/kdd/RadlinskiJ07}.
A search-engine ranker is a function that orders a collection of documents according to their relevancy to a given user search query.
By interleaving the output of two rankers and tracking on which ranker's output the user did click, we are able to get an unbiased feedback about the relative quality of these two rankers.
Given $K$ rankers, the problem of finding the best ranker is indeed a K-armed dueling bandits.

In order to obtain reproducible and comparable results, we adopted the  stochastic matrix-based experiment setup already employed by
\citet{Yue2011,zoghi2013relative,zoghi2014relative,zoghi2015mergerucb}
with both the cumulative Condorcet regret as defined by \citet{Yue/etal/12a} and the \textit{accuracy} i.e. the best arm hit-rate over the runs: $\frac{1}{N}\sum_n{\llbracket (a,b) = (i^*,i^*)\rrbracket}$.

This experimental setup uses real search engines' logs to build empirical preference matrices. The duel outcomes are then simulated on these matrices.
We used several preference matrices issued from namely: {\sc arxiv} dataset \cite{Yue2011}, {\sc letor np2004} dataset \cite{Liu07letor:benchmark}, and {\sc mslr30k} dataset.
The last dataset distinguishes three kinds of queries: informational, navigational and perfect-hit navigational \cite{MSLR30K}.
These matrices are courtesy of \citet{zoghi2014relative}'s authors.

\subsection{Empirical validation of Corollary~\ref{thm:mainTheorem2}}
\label{sec:xp1}
\begin{figure}
\begin{center}
\includegraphics[width=0.99\columnwidth]{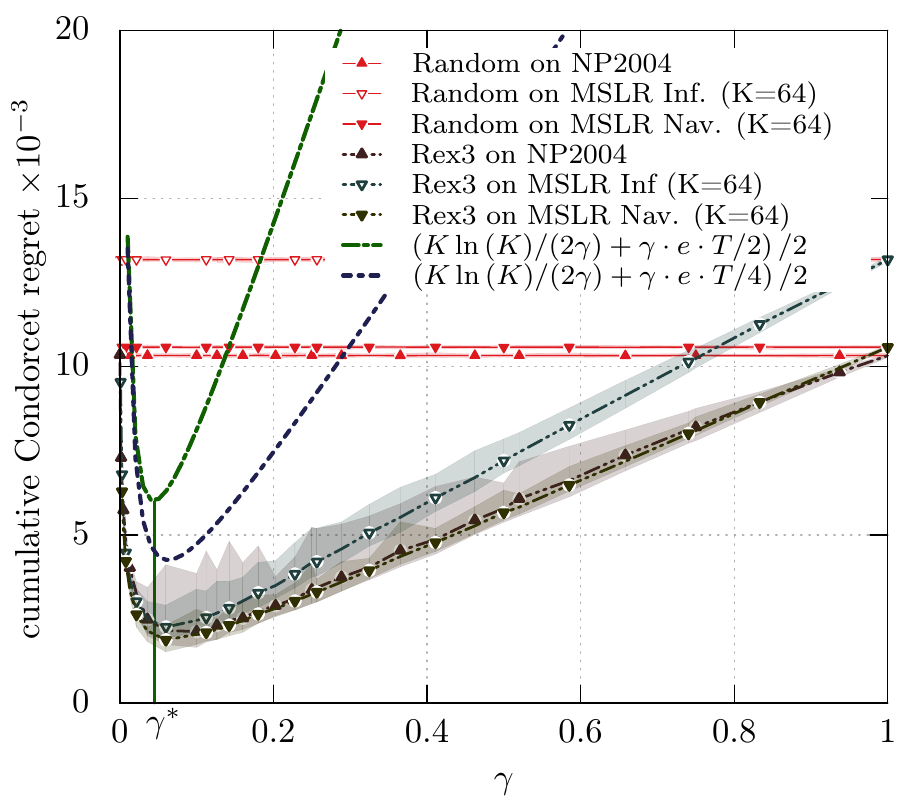}
\caption{Empirical validation of Corollary~\ref{thm:mainTheorem2}.
The colored areas around the curves show the minimal and maximal values over 100 runs.}
\label{fig:lem2sens}
\end{center}
\end{figure}

We have used {\sc letor np2004} and {\sc mslr30k} datasets (resricted to 64 rankers) to compare
the average {Condorcet regret} of $100$ runs of {\sc rex3} with $T=10^5$ to the corresponding halved\footnote{
As mentioned at the end of section~\ref{sec:previous2}, the utility-based bandit regret is indeed twice the Condorcet regret as defined in \eqref{eq:condorcetregret}.}
theoretical bounds from Corollary~\ref{thm:mainTheorem2}
for various values of $\gamma$.
The results of this experiment are sumarized in Figure~\ref{fig:lem2sens}. We plotted two theoretical curves: one with a conservative $\mathbb{G}_{max} = T/2$, and a riskier one with
$\mathbb{G}_{max} = T/4$.
This experiment illustrates the dual impact of the $\gamma$ parameter on the
exploration/exploitation tradeoff: a low value reduces both the exploration and the reactivity of the algorithm to unexpected feedbacks and a high value tends to uniformize exploration while increasing reactivity.
It also shows that the theoretical optimal $\gamma^*$ we obtain with Equation~\eqref{eq:optigamma} is a good guess even with a conservative upper-bound for $\mathbb{G}_{max}$.

\subsection{Interleave filtering simulations}
\label{sec:xp2}
For our experiments we have considered the following state of the art algorithms:
{\sc btm} \cite{Yue2011} with $\gamma=1.1$ and $\delta=1/T$ (explore-then-exploit setting),  Condorcet-{\sc savage} \cite{urvoy2013generic} with $\delta=1/T$, {\sc rucb} \cite{zoghi2013relative} with $\alpha=0.51$, and  {\sc Sparring} coupled with {\sc exp3} \cite{DBLP:conf/icml/AilonKJ14}. We also took the uniform sampling strategy {\sc Random} as a baseline.
We considered three versions of {\sc rex3}: 
%\begin{itemize}
two non-anytime versions where the optimal $\gamma^*$ is computed beforehand according to \eqref{eq:optigamma} with $\mathds{G}_ {max}$ set respectively to $T/2$ and $T/10$ and one anytime version where $\gamma^*$ is recomputed at each time step according to \eqref{eq:optigamma} \citep[see][for details about this form of ``doubling trick'']{conf/ewrl/SeldinSAA12}.
%\end{itemize}
%

A point which makes the comparison difficult is that some algorithms are anytime while others require the horizon as input.
For anytime algorithms, namely {\sc Random}, {\sc rucb} and {\sc rex3} with adaptive $\gamma$,
we displayed the average over 100 runs of the 
progressive accumulation of regret while for non-anytime algorithms, namely
{\sc btm}, {\sc csavage}, {\sc Sparring} and other versions of {\sc rex3}, we displayed the average over 50 runs of the final cumulative regret for several fixed and known horizons.
This protocol is slightly favorable to non-anytime algorithms which benefit from
more information. However, for elimination algorithms like {\sc btm} and {\sc csavage} the difference between the anytime regret and the non-anytime regret is small. For adversarial algorithms like {\sc Sparring} and {\sc rex3} the ``doubling trick'' can be applied to make them anytime: the adaptive $\gamma$ version of {\sc rex3} is an example of such a fixed-to-anytime transformation.

\begin{figure*}
\begin{center}
\begin{tabular}{cc}
\includegraphics[width=0.95\columnwidth,height=9cm]{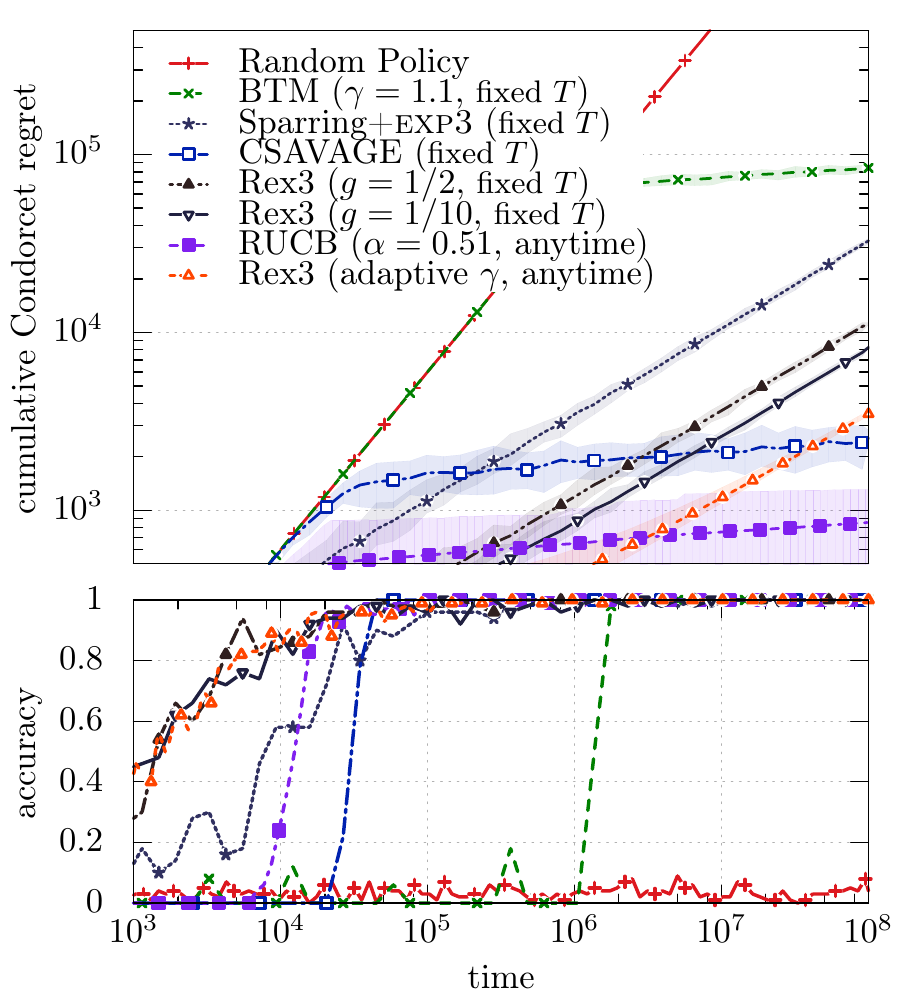}&
\includegraphics[width=0.95\columnwidth,height=9cm]{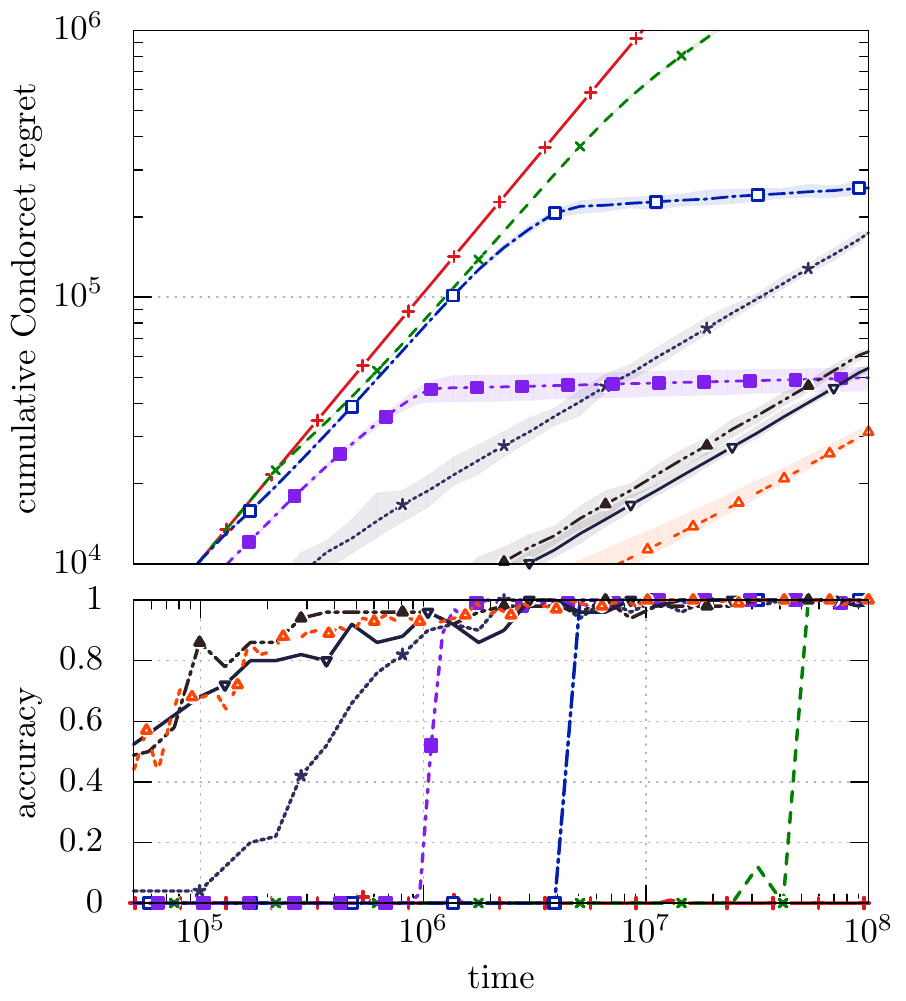}
\end{tabular}
\caption{Regret and accuracy plots averaged over 100 runs (50 runs for fixed-horizon algorithms) respectively on {\sc arxiv} dataset ($6$ rankers) and {\sc letor np2004} dataset ($64$ rankers).
{\it On regret plots, both time and regret scales are logarithmic ($\sqrt{t}$ hence appears as $t/2$). 
The colored areas around the curves show the minimal and maximal values over the runs.}
}
\label{fig:arxivnp2004}
\end{center}
\end{figure*}

\begin{figure*}
\begin{center}
\begin{tabular}{cc}
\includegraphics[width=0.45\textwidth,height=9cm]{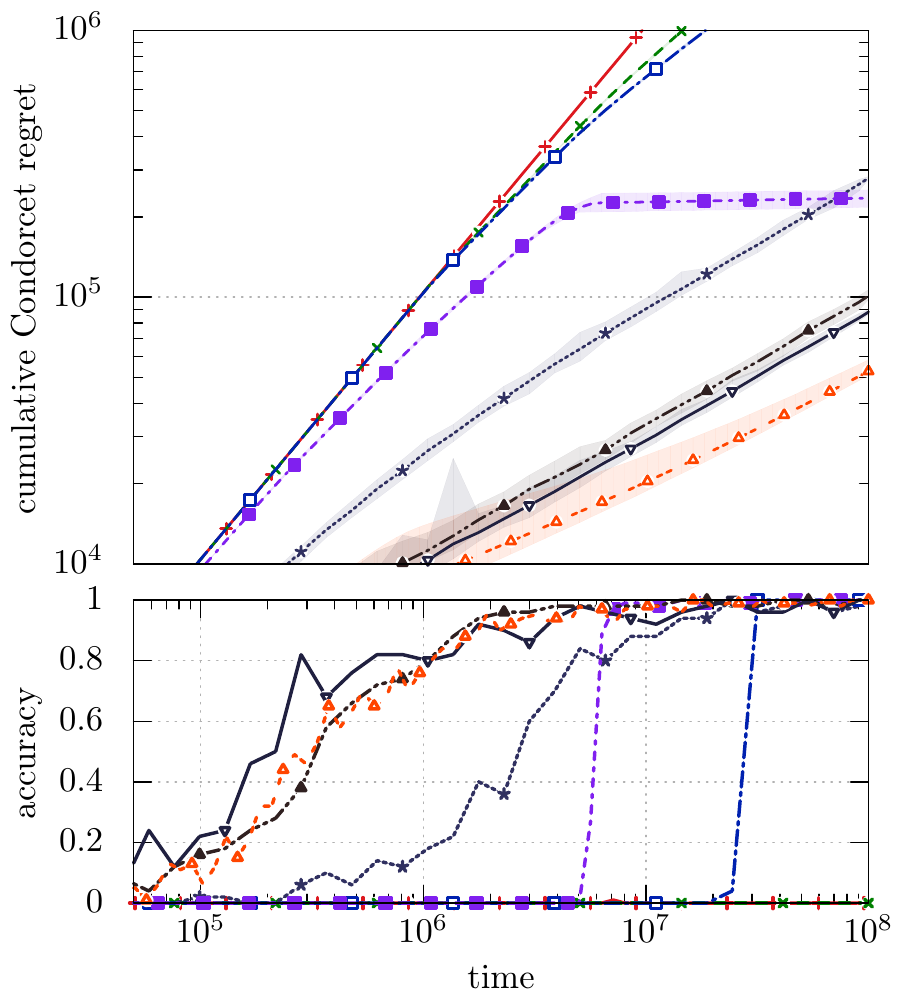}&
\includegraphics[width=0.45\textwidth,height=9cm]{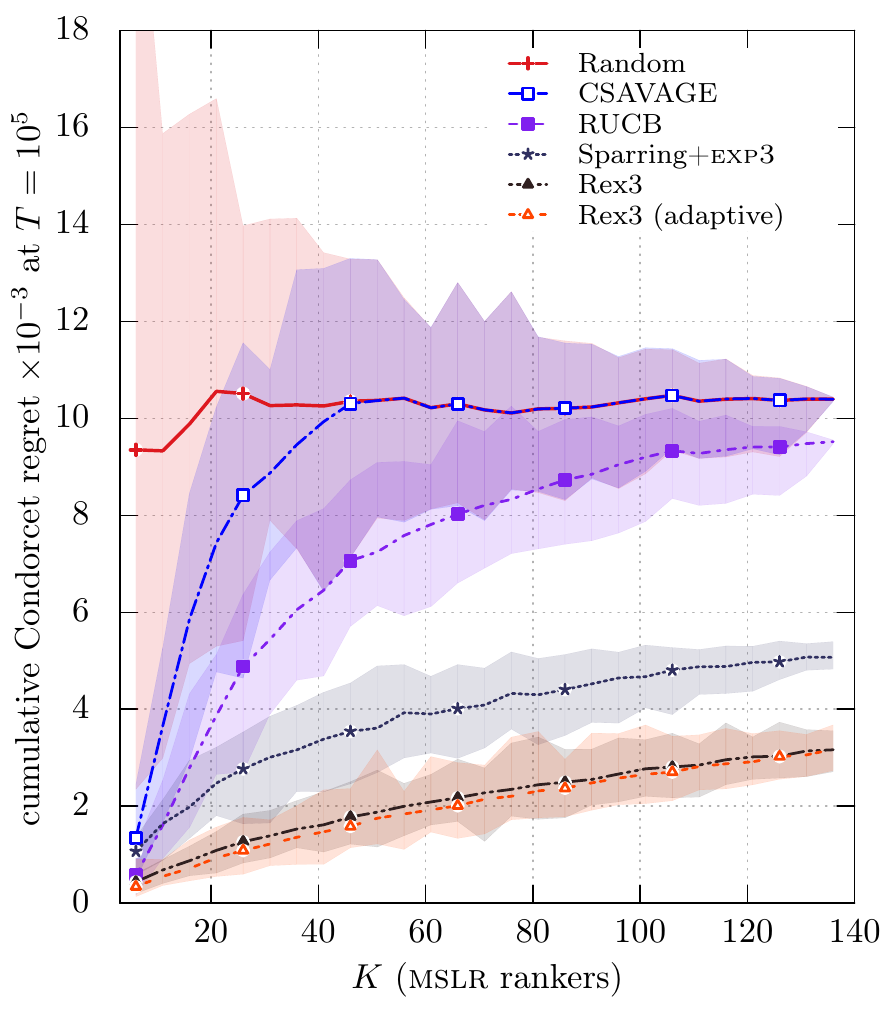}
\end{tabular}
\caption{On the left: average regret and accuracy plots on {\sc mslr30k} with navigational queries ($K=136$ rankers). On the right: same dataset, average regrets for a fixed $T=10^5$ and $K$ varying from $4$ to $136$. }
\label{fig:mslr30k}
\end{center}
\end{figure*}

The results of these experiments are summarized in Figure~\ref{fig:arxivnp2004}, 
and \ref{fig:mslr30k}. Furthermore, similar experiments are given as extended material.

As expected, the adversarial-setting algorithms {\sc Sparring} and {\sc rex3} follow an $\mathcal{O}(\sqrt{T})$ regret curve while the stochastic-setting algorithms follow an $\mathcal{O}(\ln{T})$ curve. Among the adversarial-setting algorithms, {\sc Rex3} is shown to outperform {\sc Sparring} on all datasets.
In the long run, adversarial-setting algorithms continue exploring and cannot compete in terms of regret against stochastic-setting algorithms, but the accuracy curves show that the cost of this exploration is very small. Moreover, for small horizons or high number of rankers, {\sc rex3} is extremely competitive against other algorithms.
%when $t\leq 10^6$ for the most unfavorable dataset

This difference is clearly illustrated on the left-hand side of Figure~\ref{fig:mslr30k} where we show the evolution of the expected cumulative regret at a fixed time horizon ($T=10^5$) according to the number of arms.
To obtain this plot we averaged the regret over 50 runs. For each $K$ and each run we sampled uniformly $K$ dimensions of the original $136\!\times\!136$ {\sc MSLR30k} navigational preference matrix.

%\begin{figure*}
%\begin{center}
%\begin{tabular}{ccc}
%\includegraphics[width=0.33\textwidth]{img/ARXIV2011_K6_stationary.pdf}&
%\includegraphics[width=0.33\textwidth]{img/Bvs_K20_stationary.pdf}&
%\includegraphics[width=0.33\textwidth]{img/SAVAGE_K30_stationary.pdf}\\
%%(a) & (b) & (c)
% \end{tabular}
%\caption{ Respectivley {\sc arxiv}, {\sc Bvs}, and {\sc SAVAGE}. Time and regret scales are logarithmic.}
%\label{fig:arxiv}
%\end{center}
%\end{figure*}

\section{Conclusion}
\label{sec:conclusion}

We proposed {\sc rex3}, an exponential weighing algorithm for adversarial utility-based dueling bandits. We provided both an upper and a lower bound
for its expected cumulative regret. These two bounds match the original bounds of the classical {\sc exp3} algorithm.
A thorough empirical study on several information retrieval datasets has confirmed the validity of
these theoretical results.
It also showed that {\sc rex3} and especially its anytime version with adaptive $\gamma$ are competitive solutions for dueling bandits, even when compared to stochastic-setting algorithms in a stochastic environment.

%To the best of our knowledge, there is no matrix-based formulation of the adversarial dueling bandits problem in the literature. For further works, we plan to study a pure, i.e. non-utility-based, formulation of the adversarial dueling bandits problem.
%Especially we conjecture that the lower bound regret for this general setting is quadratic in $K$.

%XXXXXXXXXXXXXXXX
\appendix

\section{Proof Sketch for Theorem~\ref{thm:mainTheorem}}
\label{sec:thmproofsk}
The general structure of the proof is similar to the one of \citep[section~3]{auer2002nonstochastic}, but, as explained before, the $\hat{c}_i(t)$ estimator we use differs from the one of {\sc exp3} because it gives an instantaneous regret estimate instead of an absolute reward estimate. As such, it may reach negative values and the $w_i(t)$ weights may decrease with time.
We only give here a sketch of proof, stressing on the differences from
\cite{auer2002nonstochastic}.
An unfamiliar reader may refer to the extended version for step-by-step details.
\begin{proof}
Let $W_t = w_1(t) + w_2(t) + \dots + w_K(t)$. As in {\sc exp3} proof we consider:
\begin{align*}
\frac{W_{t+1}}{W_t} &= \sum_{i=1}^{K} \frac{p_i(t) - \gamma/K}{1 - \gamma}
e^{(\gamma/K) \hat{c}_i(t)}
\end{align*}
The inequality $e^x \leq 1 + x + (e-2){x^2}$ is tight for $x\in [0,1]$ but it remains valid
for negative values, hence:
%\begin{align*}
%\frac{W_{t+1}}{W_t} &\leq \sum_{i=1}^{K} \frac{p_i(t) - 
%\gamma/K}{1 - \gamma}  
%\left( 1 + (\gamma/K)\hat{c}_i(t) \right) \\
%&+ \sum_{i=1}^{K} \frac{p_i(t) - \gamma/K}{1 - \gamma} \left( (e-2) (\gamma^2/K^2) \hat{c}_i(t)^2 \right)
%\end{align*}
\begin{align*}
\frac{W_{t+1}}{W_t} &\leq 1 -  \frac{\gamma^2 /K}{1- \gamma}
\left(
\underbrace{ \frac{1}{K} \sum_{i=1}^{K} \hat{c}_i(t) }_{=-M_1}
\right) \\
&+ \frac{(e-2)\gamma^2/K}{1- \gamma}
\left(
\underbrace{ \frac{1}{K} \sum_{i=1}^{K} p_i(t)\hat{c}_i(t)^2 }_{=M_2}
\right)
\end{align*}
As in {\sc exp3} we take the logarithm and sum over $t$. We get for any $j$:
\begin{align*}
\sum_{t=1}^{T} \frac{\gamma}{K}\hat{c}_j(t) - \ln(K)
&\leq \frac{\gamma^2/K}{1- \gamma} M_1 
+\frac{(e-2)\gamma^2/K}{1- \gamma} M_2
\end{align*}
%&\sum_{t=1}^{T} \frac{\gamma}{K}\hat{c}_j(t) - \ln(K) \leq 
%\sum_{i=t}^{T} \frac{\gamma^2/K^2}{1- \gamma} \frac{(p_a - p_b)(x_a - x_b)}{2p_a p_b} \\
%&+ \sum_{t=1}^{T} \frac{(e-2)\gamma^2/K^2}{1- \gamma} \frac{( p_a + p_b)(x_a - x_b)^2}{4p_a p_b}
%\end{flalign*}
By taking the expectation over the algorithm's randomization, we obtain for any $j$:
\begin{align}
&\sum_{t=1}^{T} \frac{\gamma}{K}
\underbrace{\mathds{E}_{\sim p}\hat{c}_j(t)}_{\eqref{eq:lemma1}}
  - \ln(K) \leq \notag \\
\label{eq:anyj}&\frac{\gamma^2/K}{1- \gamma} \sum_{i=t}^{T}
\underbrace{\mathds{E}_{\sim p} M_1}_{\eqref{eq:expectM1a}}
+ \frac{(e-2)\gamma^2/K}{1- \gamma} \sum_{i=t}^{T}
\underbrace{\mathds{E}_{\sim p} M_2}_{\eqref{eq:expectM2a}}
\end{align}
From Lemma~\ref{lem:expectC} we directly get the expected regret against $j$ on the left side of the inequality:
\begin{equation}
\label{eq:lemma1}\mathds{E}_{\sim p} \hat{c}_j(t) =  x_j - \mathds{E}_{\sim p}(x_a)
\end{equation}
By averaging \eqref{eq:lemma1} over the arms, we obtain:
\begin{align}
\label{eq:expectM1a}\mathds{E}_{\sim p(t)} M_1 &=
-\frac{1}{K}\sum_{i=1}^{K} \mathds{E}_{\sim p}\hat{c}_i(t)=
\mathds{E}(x_a) - \frac{1}{K}\sum_{i=1}^{K}x_i
\end{align}
The following result is detailled in the extended version:
\begin{align}
%\mathds{E}_{\sim p(t)} M_2 &=
%\frac{1}{2}\mathds{E}(x_a^2) - \mathds{E}(x_a)\frac{1}{K}\sum_{i=1}^{K}x_i + \frac{1}{2K}\sum_{i=1}^{K}x_i^2 \notag \\
\label{eq:expectM2a}\mathds{E}_{\sim p(t)} M_2&\leq \frac{1}{2}\mathds{E}(x_a) + \frac{1}{2K}\sum_{i=1}^{K}x_i%\quad{}\text{ as } x_i\in[0,1]
\end{align}
From Lemma~\ref{lem:expectC}, \eqref{eq:expectM1a}, \eqref{eq:expectM2a}, and
by definition of $\mathbb{G}_{max}$, $\mathds{E}\mathbb{G}_{alg}$, and
$\mathds{E}\mathbb{G}_{unif}$, the inequality \eqref{eq:anyj} rewrites into:
\begin{align*}
&\mathbb{G}_{max} - \mathds{E}\mathbb{G}_{alg} - \frac{K\ln{}K}{\gamma} \leq
\frac{\gamma}{1- \gamma} 
\left( \mathds{E}\mathbb{G}_{alg} - \mathds{E}\mathbb{G}_{unif} \right) \\
&+\frac{(e\!-\!2) \gamma}{2(1- \gamma)}
\left(\mathds{E}\mathbb{G}_{alg}   + \mathds{E}\mathbb{G}_{unif} 
\right)
\end{align*}
Assuming $\gamma\leq \demi{}$ we finally obtain:
\begin{align*}
\label{eq:final1}&\mathbb{G}_{max} - \mathds{E}\mathbb{G}_{alg}\leq
\frac{K\ln{}K}{\gamma} 
+\gamma\left({e}\mathds{E}\mathbb{G}_{alg}
-\left(4\!-\!e\right)\mathds{E}\mathbb{G}_{unif}\right)
\end{align*}
\end{proof}

% Acknowledgements should only appear in the accepted version. 
\section*{Acknowledgments} 

We would like to thank the reviewers for their constructive comments and Masrour Zoghi who kindly sent us his experiment datasets.

\bibliographystyle{apalike}
\bibliography{references}

%%%%%%%%%%%%%%%%%%%%%%%%%%%%%%%%%%
%%%%%%%%%%%%%%%%%%%%%%%%%%%%%%%%%%
\iftoggle{long-version}{
\onecolumn
\newpage
\icmltitle{An Exponential Weighing Algorithm for Adversarial Utility-based Dueling Bandits\iftoggle{long-version}{ (supplementary material)}{}}

%\tableofcontents

\section{Detailed Proof of Theorem~\ref{thm:mainTheorem}}
\label{sec:thmproof}
For better readabilty, we simply write $a,b$ instead of $a_t,b_t$ when referring of the arms chosen by the algorithm. We also frequently drop the time indices for $p_{.}(t)$ and $x_{.}(t)$. 
See Table~\ref{tab:notation} for an exhaustive list of notations.
  
\begin{proof}
Let $W_t = w_1(t) + w_2(t) + \dots + w_K(t)$
\begin{equation*}
\frac{W_{t+1}}{W_t} = \sum_{i=1}^{K} \frac{w_i(t+1)}{W_t}
= \sum_{i=1}^{K} \frac{w_i(t)}{W_t}
e^{(\gamma/K)\hat{c}_i(t)} 
\end{equation*}
By substituting the values of $\frac{w_i(t)}{W_t}$, we get:
\begin{align*}
\frac{W_{t+1}}{W_t} &= \sum_{i=1}^{K} \frac{p_i(t) - \gamma/K}{1 - \gamma}
e^{(\gamma/K) \hat{c}_i(t)}
\end{align*}
Using the inequality $e^x \leq 1 + x + (e-2){x^2}$ for $x \leq 1$, we get:
\begin{align}
\frac{W_{t+1}}{W_t} &\leq \sum_{i=1}^{K} \frac{p_i(t) - 
\gamma/K}{1 - \gamma}  
\left( 1 + (\gamma/K)\hat{c}_i(t) \right) 
+ \sum_{i=1}^{K} \frac{p_i(t) - \gamma/K}{1 - \gamma} \left( (e-2) (\gamma^2/K^2) \hat{c}_i(t)^2 \right)\notag \\
&\leq \underbrace{\sum_{i=1}^{K} \frac{p_i(t) - \gamma/K}{1 - \gamma}}_{=1}
+  \frac{\gamma /K}{1- \gamma}  \left( \underbrace{ \sum_{i=1}^{K} p_i(t) \hat{c}_i(t)}_{=0 \text{ see } \eqref{eq:sumpc1}} - \frac{\gamma}{K} 
{\sum_{i=1}^{K} \hat{c}_i(t)} \right)
+ \frac{(e-2) (\gamma^2/K^2)}{1- \gamma}
{\sum_{i=1}^{K} p_i(t)\hat{c}_i(t)^2}\notag \\
&\leq 1
+  \frac{\gamma /K}{1- \gamma}  \left( 0 - \frac{\gamma}{K} {\sum_{i=1}^{K} \hat{c}_i(t)} \right)
+ \frac{(e-2) (\gamma^2/K^2)}{1- \gamma}  {\sum_{i=1}^{K} p_i(t)\hat{c}_i(t)^2}\notag \\
\label{eq:M1M2}&\leq 1
-  \frac{\gamma^2 /K}{1- \gamma}  \left(\underbrace{\frac{1}{K}\sum_{i=1}^{K} \hat{c}_i(t)}_{=-M_1} \right)
+ \frac{(e-2) (\gamma^2/K)}{1- \gamma}  \underbrace{\frac{1}{K}\sum_{i=1}^{K} p_i(t)\hat{c}_i(t)^2}_{=M_2}
\end{align}
\begin{align}
\sum_{i=1}^{K} p_i(t) \hat{c}_i(t) &= \sum_{i=1}^{K} p_i(t) \left( \llbracket{}i = a\rrbracket{} \frac{x_a - x_b}{2p_a} \right)
+ \sum_{i=1}^{K} p_i(t) \left( \llbracket{}i = b \rrbracket{} \frac{x_b - x_a}{2p_b} \right)\notag \\
\label{eq:sumpc1}&= \frac{x_a - x_b}{2} + \frac{x_b - x_a}{2}= 0
\end{align}
%
%
%
%\begin{align}
%M_1 &\doteq{}-\frac{1}{K}\sum_{i=1}^{K} \hat{c}_i(t)
%=-\sum_{i=1}^{K} \left( \llbracket{}i = a \rrbracket{} \frac{x_a - x_b}{2Kp_a} + \llbracket{} i = b \rrbracket{} \frac{x_b - x_a}{2Kp_b} \right)
%&= \frac{x_a - x_b}{2p_a} + \frac{x_b - x_a}{2p_b}\notag \\
%= \frac{p_b(x_b - x_a)}{2Kp_a p_b} + \frac{p_a(x_a - x_b)}{2Kp_a p_b}\notag\\
%\label{eq:sumc1} &= \frac{(p_a - p_b)(x_a - x_b)}{2Kp_a p_b} 
%\end{align}
%
%
%
%\begin{align}
%\label{eq:sumpc2}M_2 &\doteq {}\frac{1}{K}\sum_{i=1}^{K} p_i(t)\hat{c}_i(t)^2=\frac{( p_a + p_b)(x_a - x_b)^2}{4Kp_a p_b} 
%\end{align} 
%The proof of \eqref{eq:sumpc2} is detailed on Appendix~\ref{proof:eq:sumpc2}.

From \eqref{eq:M1M2} and \eqref{eq:sumpc1}, we obtain:
\begin{align*}
\frac{W_{t+1}}{W_t} &
%\leq 1 +  \frac{\gamma /K}{1- \gamma} \left( 0 - \frac{\gamma}{K} \frac{(p_b - p_a)(x_a - x_b)}{2p_a p_b} \right) \\ &+ \frac{(e-2) (\gamma^2/K^2)}{1- \gamma} \frac{( p_a + p_b)(x_a - x_b)^2}{4p_a p_b} \\
%&\leq 1 + \frac{\gamma^2/K^2}{1- \gamma}
% \frac{(p_a - p_b)(x_a - x_b)}{2p_a p_b}  \\ &+ \frac{(e-2) (\gamma^2/K^2)}{1- \gamma} \frac{( p_a + p_b)(x_a - x_b)^2}{4p_a p_b} 
\leq 1 + \frac{\gamma^2/K}{1- \gamma}
M_1+
\frac{(e-2) (\gamma^2/K)}{1- \gamma}
M_2
\end{align*}
Taking logarithms and using the inequality $1+x \leq e^x$
\begin{align*}
%\ln \frac{W_{t+1}}{W_t} &\leq \frac{\gamma^2/K^2}{1- \gamma} \frac{(p_a - p_b)(x_a - x_b)}{2p_a p_b}  \\ &+ \frac{(e-2) (\gamma^2/K^2)}{1- \gamma} \frac{( p_a + p_b)(x_a - x_b)^2}{4p_a p_b} 
\ln \frac{W_{t+1}}{W_t} &\leq \frac{\gamma^2/K}{1- \gamma} M_1 +
\frac{(e-2) (\gamma^2/K)}{1- \gamma} 
M_2
\end{align*}
Summing over $t$, we get:
\begin{align}
\label{eq:logWT}\ln \frac{W_{T+1}}{W_1} &\leq
\frac{\gamma^2/K}{1- \gamma} M_1 +
\frac{(e-2) (\gamma^2/K)}{1- \gamma} 
M_2
\end{align}

For any arm $j$ we have:
\begin{equation}
\label{eq:Wanyj}
\sum_{t=1}^{T} \frac{\gamma}{K}\hat{c}_j(t) - \ln(K) \leq \ln \frac{W_{T+1}}{W_1}
\end{equation}
The proof of the above inequality is given in appendix~\ref{proof:eq:Wanyj}.
By combining \eqref{eq:logWT} and \eqref{eq:Wanyj} , we get:
\begin{align}
\label{eq:logWT2}
\sum_{t=1}^{T} \frac{\gamma}{K}\hat{c}_j(t) - \ln(K)
&\leq \frac{\gamma^2/K}{1- \gamma} M_1 +
\frac{(e-2)\gamma^2/K}{1- \gamma} 
M_2
\end{align}
%&\sum_{t=1}^{T} \frac{\gamma}{K}\hat{c}_j(t) - \ln(K) \leq 
%\sum_{i=t}^{T} \frac{\gamma^2/K^2}{1- \gamma} \frac{(p_a - p_b)(x_a - x_b)}{2p_a p_b} \\
%&+ \sum_{t=1}^{T} \frac{(e-2)\gamma^2/K^2}{1- \gamma} \frac{( p_a + p_b)(x_a - x_b)^2}{4p_a p_b}
%\end{flalign*}
Taking the expectation over the algorithm's randomization, we obtain:
\begin{align*}
\sum_{t=1}^{T} \frac{\gamma}{K}
\underbrace{\mathds{E}_{\sim p}\hat{c}_j(t)}_{\text{Lemma~\ref{lem:expectC}}}
  - \ln(K)  &\leq \frac{\gamma^2/K}{1- \gamma} \sum_{i=t}^{T}
\underbrace{\mathds{E}_{\sim p} M_1}_{\text{see }\eqref{eq:expectM1}}
+ \frac{(e-2) (\gamma^2/K)}{1- \gamma} \sum_{i=t}^{T}
\underbrace{\mathds{E}_{\sim p} M_2}_{\text{see } \eqref{eq:expectM2}}
\end{align*}
From Lemma~\ref{lem:expectC} which proof is detailed in appendix~\ref{sec:expectC}, we have:
\begin{equation}
\label{eq:eqLem1}\mathds{E}_{\sim p} \hat{c}_j(t) =  x_j - \mathds{E}_{\sim p}(x_a)
\end{equation}
By averaging \eqref{eq:eqLem1} over the arms, we obtain:
\begin{align}
\mathds{E}_{\sim p(t)} M_1 &=
\mathds{E}_{\sim p}\left(-\frac{1}{K}\sum_{i=1}^{K} \hat{c}_i(t)\right)
=-\frac{1}{K}\sum_{i=1}^{K} \mathds{E}_{\sim p}\hat{c}_i(t)
\label{eq:expectM1}=
\mathds{E}(x_a) -  \frac{1}{K}\sum_{i=1}^{K}x_i
\end{align}
The following result is detailled in appendix~\ref{sec:expectM2}:
\begin{align}
\mathds{E}_{\sim p(t)} M_2 &=
\mathds{E}_{\sim p(t)} \left(\frac{( p_a + p_b)(x_a - x_b)^2}{4Kp_a p_b}\right)
\label{eq:expectM2}=\frac{1}{2}\mathds{E}(x_a^2) - \mathds{E}(x_a)\frac{1}{K}\sum_{i=1}^{K}x_i + \frac{1}{2K}\sum_{i=1}^{K}x_i^2 \notag \\
&\leq \frac{1}{2}\mathds{E}(x_a) - \mathds{E}(x_a)\frac{1}{K}\sum_{i=1}^{K}x_i + \frac{1}{2K}\sum_{i=1}^{K}x_i \quad\quad\text{ as } \forall i,\ x_i\in [0,1] 
\end{align}
From Lemma~\ref{lem:expectC}, \eqref{eq:expectM1}, and \eqref{eq:expectM2}, we get for any $j$:
\begin{align}
&\frac{\gamma}{K}\left( \sum_{t=1}^{T}x_j - \sum_{t=1}^{T}\mathds{E}(x_a)\right) - \ln(K) \leq
\frac{\gamma^2/K}{1- \gamma} \sum_{i=t}^{T}
\left(\mathds{E}(x_a) -  \frac{1}{K}\sum_{i=1}^{K}x_i\right)\notag\\
\label{eq:expectR1}&+ \frac{(e\!-\!2) \gamma^2/K}{2(1- \gamma)} \sum_{i=t}^{T}
\left(\mathds{E}(x_a^2) - 2\mathds{E}(x_a)\frac{1}{K}\sum_{i=1}^{K}x_i + \frac{1}{K}\sum_{i=1}^{K}x_i^2\right)
\end{align}

By definition, $\mathbb{G}_{max}=\max_j\sum_{t=1}^{T}x_j$,
$\quad \mathds{E}\mathbb{G}_{alg}=\sum_{t=1}^{T} \mathds{E}_{\sim{}p(t)}(x_a)$,
and $\quad\mathds{E}\mathbb{G}_{unif}=\sum_{t=1}^{T} \frac{1}{K}\sum_{i=1}^Kx_i$.
We can hence rewrite Equation~\eqref{eq:expectR1} into:
\begin{align}
&\mathbb{G}_{max} - \mathds{E}\mathbb{G}_{alg} - \frac{K\ln{}K}{\gamma} \leq
\frac{\gamma}{1- \gamma} 
\left( \mathds{E}\mathbb{G}_{alg} - \mathds{E}\mathbb{G}_{unif} \right)
+\frac{(e\!-\!2) \gamma}{2(1- \gamma)}
\left(\mathds{E}\mathbb{G}_{alg}   + \mathds{E}\mathbb{G}_{unif} 
-2 \sum_{t=1}^T\sum_{i=1}^K \frac{x_i}{K} \mathds{E}(x_a)\right)
\end{align}
Let $\varepsilon$ be such that $\forall i,t\ \varepsilon\leq{}x_i(t)$ then:
\begin{align}
&\mathbb{G}_{max} - \mathds{E}\mathbb{G}_{alg}  \leq  \frac{K\ln{}K}{\gamma}
+\frac{e\gamma}{2(1-\gamma)}\mathds{E}\mathbb{G}_{alg}
-\frac{(4\!-\!e+(e\!-\!2)\varepsilon)\gamma}{2(1-\gamma)}
\mathds{E}\mathbb{G}_{unif}
\end{align}
Assuming $\gamma\leq \demi{}$:
\begin{align}
\label{eq:final1}&\mathbb{G}_{max} - \mathds{E}\mathbb{G}_{alg}\leq
\frac{K\ln{}K}{\gamma} 
+\gamma\left[{e}\mathds{E}\mathbb{G}_{alg}
-\left(4\!-\!e+(e\!-\!2)\varepsilon\right)\mathds{E}\mathbb{G}_{unif}\right]
\end{align}

%XXXXXXX

%\begin{align*}
%(\gamma/K) \left( \sum_{t=1}^{T}x_i - \sum_{t=1}^{T}\mathds{E}_{\sim P}(x_a)\right) - \ln(K) &\leq \frac{\gamma^2/K^2}{1- \gamma} \sum_{i=t}^{T} \sum_{a=1}^{K} \sum_{b=1}^{K} \frac{(p_a - p_b)(x_a - x_b)}{2}
%\end{align*}

\end{proof}

\subsection{Proof of eq. \eqref{eq:Wanyj}}
\label{proof:eq:Wanyj}
For any $j$ we have:  
%\begin{align*}
$\qquad{}W_{T+1} = \sum_{i=1}^{K}w_i(T+1)
\quad{}\geq\quad{} w_j(T+1)$.
%\end{align*}

Hence:
\begin{align*}
W_{T+1} \quad{}&\geq\quad{} w_j(T) e^{(\gamma/K) (\hat{c}_j(T))}
= w_j(T-1) e^{(\gamma/K) (\hat{c}_j(T-1))} e^{(\gamma/K) (\hat{c}_j(T))}
= w_j(1)\prod_{t=1}^{T} e^{(\gamma/K) (\hat{c}_j(t))},\quad \text{ and}\\
%W_{T+1}\quad{}&\geq\quad{} e^{ \sum_{t=1}^{T} \frac{\gamma}{K}\hat{c}_j(t)} \\
\ln W_{T+1}\quad{}&\geq\quad{} \ln w_j(1)\quad{}+\quad \sum_{t=1}^{T} \frac{\gamma}{K}\hat{c}_j(t)\\
\end{align*}
Since $w_j(1) = 1$ for any $j$, it turns out that:
$
\sum_{t=1}^{T} \frac{\gamma}{K}\hat{c}_j(t) - \ln(K) \leq \ln (\frac{W_{T+1}}{W_1})$
\qed

\subsection{Proof of Lemma~\ref{lem:expectC}}
\label{sec:expectC}
\begin{align*}
 \hat{c}_i(t) &= \llbracket{}i = a_t\rrbracket{} \frac{\psi(x_{a_t} - x_{b_t})}{2p_{a_t}(t)} + \llbracket{}i = b_t\rrbracket{} \frac{\psi(x_{b_t} - x_{a_t})}{2p_{b_t}(t)} \\[+5pt]
 \mathds{E}_{(a,b)\sim{}p(t)}\hat{c}_i(t) &= \sum_{j=1}^{K} \sum_{k=1}^{K} p_j(t) p_k(t) \left( \llbracket{}i = j\rrbracket{}
 \frac{\psi(x_j - x_k)}{2p_j} + \llbracket{}i = k\rrbracket{}
 \frac{\psi(x_k - x_j)}{2p_k} \right) \\
&= \sum_{j=1}^{K} \sum_{k=1}^{K} p_j p_k \llbracket{}i = j\rrbracket{} \frac{\psi(x_j - x_k)}{2p_j} + \sum_{j=1}^{K} \sum_{k=1}^{K} p_j p_k \llbracket{}i = k\rrbracket{}
\frac{\psi(x_k - x_j)}{2p_k} \\
 &= \frac{1}{2} \sum_{k=1}^{K} p_k \psi(x_i - x_k) + \frac{1}{2} \sum_{j=1}^{K} p_j \psi(x_i - x_j)\\
 % \quad{}=\quad{} \frac{1}{2} \left( \sum_{b=1}^{K} p_b x_i - \sum_{b=1}^{K} p_b x_b + \sum_{a=1}^{K} p_a x_i - \sum_{a=1}^{K} p_a x_a   \right) \\
& = \mathds{E}_{a\sim{}p} \psi(x_i - x_a)
\end{align*}
If $\psi(x) = x$ it simplifies into:
\begin{align*}
\mathds{E}_{(a,b)\sim{}p(t)} \hat{c}_i(t) &=
\quad{} x_i - \mathds{E}_{a\sim p(t)}x_a
\end{align*}\qed

%\subsection{Proof of eq. \eqref{eq:expectM1a} and \eqref{eq:expectM1}}
%\label{sec:expectM1}
%%To simplify notation we drop the $t$ indices on $x_{.}(t)$ and $p_{.}(t)$.
%\begin{align*}
%\mathds{E}_{(a,b)\sim p(t)} M_1 &= \mathds{E}_{(a,b)\sim p(t)}\left( \frac{(p_a - p_b)(x_a - x_b)}{2K p_a p_b} \right)\\
%%
%&= \sum_{i=1}^{K}
%\sum_{j=1}^{K} p_i p_j \frac{(p_i - p_j)(x_i - x_j)}{2K p_i p_i}
%\quad{}=\quad{} \frac{1}{2K} \sum_{i=1}^{K}
%\sum_{j=1}^{K} (p_i - p_j)(x_i - x_j) \notag \\
%%
%&=  \frac{1}{2K} \left(\sum_{i=1}^{K} \sum_{j=1}^{K} p_i x_i  - \sum_{i=1}^{K} \sum_{j=1}^{K} p_i x_j - \sum_{i=1}^{K} \sum_{j=1}^{K} p_j x_i + \sum_{i=1}^{K} \sum_{j=1}^{K} p_j x_j\right) \notag \\
%%
%&= \frac{1}{2K} \left( K \sum_{i=1}^{K}p_i x_i -  \sum_{i=1}^{K} p_i \sum_{j=1}^{K}x_j - \sum_{j=1}^{K} p_j \sum_{i=1}^{K} x_i + K \sum_{j=1}^{K} p_j x_j\right) \notag\\
%%
%&= \frac{1}{2K}\left(2K \mathds{E}_{\sim{}p}(x_a) -  2 \sum_{i=1}^{K}x_i\right) \notag \\
%&= \mathds{E}_{\sim{}p}(x_a) -  \frac{1}{K}\sum_{i=1}^{K}x_i
%\end{align*}\qed

\subsection{Proof of eq. \eqref{eq:expectM2a} and \eqref{eq:expectM2}}
\label{sec:expectM2}
%To simplify notation we drop the $t$ indices on $x_{.}(t)$ and $p_{.}(t)$.

%\section{Proof of eq. \eqref{eq:sumpc2}}
%\label{proof:eq:sumpc2}
%To simplify notation we write resp. $a,b$ for $a_t,b_t$ and drop the $(t)$ indices on $x_{.}(t)$ and $p_{.}(t)$.
\begin{align*}
M_2&=\frac{1}{K}\sum_{i=1}^{K} p_i(t) \hat{c}_i(t)^2 \\ 
&= \frac{1}{K}\sum_{i=1}^{K} p_i\left( \llbracket{}i = a\rrbracket{} \frac{x_a - x_b}{2p_a} + \llbracket{}i = b\rrbracket{} \frac{x_b - x_a}{2p_b} \right)^2 \\
&= \frac{1}{K}\sum_{i=1}^{K} \left( p_i \llbracket{}i = a\rrbracket{} \frac{(x_a - x_b)^2}{4p_a^2} + p_i(t)\llbracket{}i = b\rrbracket{} \frac{(x_b - x_a)^2}{4p_b^2} + \underbrace{2p_i\llbracket{}i = a\rrbracket{} \llbracket{}i = b\rrbracket{} \frac{x_a - x_b}{2p_a} \frac{x_b - x_a}{2p_b}}_{=0} \right) \\
&=\sum_{i=1}^{K} p_i\llbracket{}i = a\rrbracket{} \frac{(x_a - x_b)^2}{4Kp_a^2} + \sum_{i=1}^{K} p_i\llbracket{}i = b\rrbracket{} \frac{(x_b - x_a)^2}{4Kp_b^2}
\quad{}=\quad{} \frac{(x_a - x_b)^2}{4Kp_a} + \frac{(x_b - x_a)^2}{4Kp_b} \\ 
%&= \frac{p_b(x_a - x_b)^2}{4p_a p_b} + \frac{p_a(x_b - x_a)^2}{4p_b p_a} \\
&= \frac{( p_a + p_b)(x_a - x_b)^2}{4Kp_a p_b}
\end{align*}

\begin{align*}
\mathds{E}_{(a,b)\sim p(t)} M_2 &=
\mathds{E}_{(a,b)\sim p(t)}\left(\frac{(p_a + p_b)(x_a - x_b)^2}{4K p_a p_b}\right)\\
&=\sum_{i=1}^{K} \sum_{j=1}^{K} p_i p_j \frac{(p_i + p_j)(x_i - x_j)^2}{4K p_i p_j}
\quad{}=\quad{} \frac{1}{4K}
\sum_{i=1}^{K} \sum_{j=1}^{K} {(p_i + p_j)(x_i - x_j)^2} \\
&=\frac{1}{4K} \sum_{i=1}^{K} \sum_{j=1}^{K} (p_i + p_j) (x_i^2 - 2x_ix_j + x_j^2)\notag \\
&= \frac{1}{4K}\left(\sum_{i=1}^{K} \sum_{j=1}^{K}p_ix_i^2 - 2\sum_{i=1}^{K} \sum_{j=1}^{K}p_ix_ix_j + \sum_{i=1}^{K} \sum_{j=1}^{K}p_ix_j^2\right.  
\left. + \sum_{i=1}^{K} \sum_{j=1}^{K}p_jx_i^2 - 2\sum_{i=1}^{K} \sum_{j=1}^{K}p_jx_ix_j + \sum_{i=1}^{K} \sum_{j=1}^{K}p_jx_j^2\right) \notag \\
&= \frac{1}{4K}\left(2K\mathds{E}(x_a^2) - 4\mathds{E}(x_a)\sum_{i=1}^{K}x_i + 2\sum_{i=1}^{K}x_i^2\right) \notag \\
&= \frac{1}{2}\left(\mathds{E}_{a\sim{}p(t)}(x_a^2) - 
2\mathds{E}_{a\sim{}p(t)}(x_a)\frac{1}{K}\sum_{i=1}^{K}x_i + \frac{1}{K}\sum_{i=1}^{K}x_i^2\right)
\end{align*}
For any $x\in [0,1]$, $x^2<x$, hence:
\begin{align*}
\mathds{E}_{(a,b)\sim p(t)} M_2 &\leq
\frac{1}{2}\left(\mathds{E}_{a\sim{}p(t)}(x_a) - 
2\mathds{E}_{a\sim{}p(t)}(x_a)\frac{1}{K}\sum_{i=1}^{K}x_i + \frac{1}{K}\sum_{i=1}^{K}x_i\right)
\end{align*}\qed

\begin{table}[h]
\begin{center}
\caption{Notation table}
\label{tab:notation}
\begin{tabular}{c@{\quad}l@{\quad}l}
\bf Notation & \bf Description & \bf Remarks \\ \Xhline{4\arrayrulewidth}
$K$ & Number of arms & \\ \hline
$t$ & Time index & \\ \hline
$T$ & Time horizon & \\ \hline
$R_T$ & Cumulative regret after time $T$ & \\  \hline
$\mathds{E}_{\sim{}\pi}(\ldots)$ & Expectation according to $\pi$ &\\ \hline
$\gamma$ & {\sc rex3} exploration/exploitation parameter & $\gamma\in[0,1/2]$ \\ 
& May also refer to {\sc btm} \cite{Yue2011} parameter. & In that case: $\gamma\geq 1$ \\ \hline
$\gamma^*$ & Optimal $\gamma$ as defined in \eqref{eq:optigamma} &  \\ \hline
$\alpha$ & Input parameter in {\sc rucb} \cite{zoghi2013relative}  & $\alpha > \frac{1}{2}$ \\ \hline
$a_t,b_t$ or $a,b$ & Pulled arms at time $t$ &\\ \hline
$i,j,k$ & Reserved for (fixed) arms indices &\\ \hline
$P$ & Preference matrix & \\  \hline
$P_{i,j}$ & The probability of arm $i$ winning against arm $j$ &  \\ \hline 
$r'_{a,b}$ & Condorcet regret \eqref{eq:condorcetregret} incurred when arms $a$ and $b$ are pulled & \\ \hline
$x_{a}(t)$ or $x_{a}$ & Reward/utility of arm $a_t$ drawn at time instant $t$ & \\ \hline
$\mathbf{x}(t)$ & Vector containing rewards of all $K$ arms at time $t$  & $\mathbf{x}(t)=(x_1(t),\ldots,x_K(t))$\\ \hline
$v_i$ & Probability distribution associated with arm $i$ & For stochastic models \\ \hline
$\mu_i$ & Mean of probability distribution $v_i$  &$\mu_i=\mathds{E}_{\sim{}v_i}x_i$ \\ \hline
$\boldsymbol{\mu}$ & Mean reward vector  & $\boldsymbol{\mu}=(\mu_1,\ldots,\mu_K)$ \\ \hline
$r_{a,b}(t)$ & Bandit regret \eqref{eq:banditregret} incurred at time instant $t$ when arms $a$ and $b$ are pulled. & On stochastic settings: \\
& & $\mathds{E}r_{a,b}(t)=2r'_{a,b}$\\ \hline
$\llbracket{} \ldots \rrbracket{}$ & Indicator function & \\ \hline
$i^*$ & Optimal arm index: Copeland/Condorcet winner of preference matrix &  $i^*\in\argmax\sum_{j=1}^K \llbracket{}P_{i,j}>\demi\rrbracket\ $\\ 
& For utility based  statement: $i^*\in\argmax \mu_i$ &   \\ \hline
$\psi$ & Feedback transfer function & $\psi(x_a-x_b)=x_a-x_b$ \\ 
 &  & or $\psi(x_a-x_b)=\llbracket{}x_a\geq{}x_b\rrbracket{}$ \\ \hline
$w_i(t)$ or $w_i$ & {\sc exp3} weight of arm $i$ at time $t$ & \\ \hline
$W_t$ & Sum of weights at time $t$ & $W_t=\sum_iw_i(t)$\\ \hline
$p_i(t)$ &  {\sc exp3} probablity that arm $i$ is pulled at time $t$& \\ \hline
$\mathbf{p}(t)$ & Probablity distribution containing & $\mathbf{p}(t)=(p_1(t), \dots, p_K(t))$\\ \hline
$\hat{c}_i(t)$ & {\sc exp3} regret estimator as defined in \eqref{eq:chat} & \\ \hline
$\mathbb{G}_{max}$ & Maximum possible gain for a single-arm strategy & \\ \hline
$\mathbb{G}_{alg}$ & Gain earned due to the strategy of \textit{alg}  &   \\ \hline
$\mathds{E}\mathbb{G}_{unif}$ & Gain earned due to the uniform strategy & $\mathds{E}\mathbb{G}_{unif}=\frac{1}{K}\sum_t\sum_i x_i(t)$\\ \hline
$\mathbb{G}_{min}$ & Minimum possible gain for a single-arm strategy & \\ \hline
$M_1, M_2$ & See \eqref{eq:M1M2} & \\ \hline
$\mathcal{L}$ & Partial monitoring Loss matrix & \\ \hline
$\mathcal{H}$ & Partial monitoring Feedback matrix & \\ \hline
$\mathcal{K}$ & {\sc FeedExp3} transfer matrix & \\ \hline
\end{tabular}
\end{center}
\end{table}

\section{Further Experiments}
\label{sec:xpbis}

We give here the simulation results that could not fit on the core article:
Figure~\ref{fig:np2004bis} gives results for smaller number of rankers on {\sc np2004} dataset and Figure~\ref{fig:mslr30kbis} complete the experiments on {\sc MSLR30k} dataset.
On Figure~\ref{fig:sparringUCB} we added an experiment we made with Sparring coupled with
{\sc ucb}.
We also considered two artificial matrices: {\sc savage} and {\sc Bvs}.
The $30\!\times{}\!30$ {\sc savage} matrix, defined by $P_{i,j}=\demi{}+j/(2K)$ for $i<j$ as described in \cite{urvoy2013generic}.
The $20\!\times{}\!20$ {\sc Bvs} matrix is defined by: $P_{1,j}=0.51$ for any $j>1$ and $P_{i,j}=1$ for any $1<i<j$.
Its Condorcet winner has a low Borda score ($9.69$ against $18.49$ for the Borda winner) which makes it difficult for algorithms to find the real Condorcet winner.
These experiments results are sumarized in Figure~\ref{fig:artificial}. 
The preference matrices we used and their properties are summarized on Table~\ref{tab:matrices}.

We conclude these experiments by a non-stationnary utility-based dueling bandit simulation where the expected reward gap $\Delta(t)$ between the best arm and the others is set in order to decieve stochastic algorithms
(see Figure~\ref{fig:adv}).

%\begin{center}
%\includegraphics[width=0.4\textwidth]{img/key.pdf}
%\end{center}
\begin{figure*}
\begin{center}
\begin{tabular}{cc}
 \includegraphics[width=0.45\textwidth]{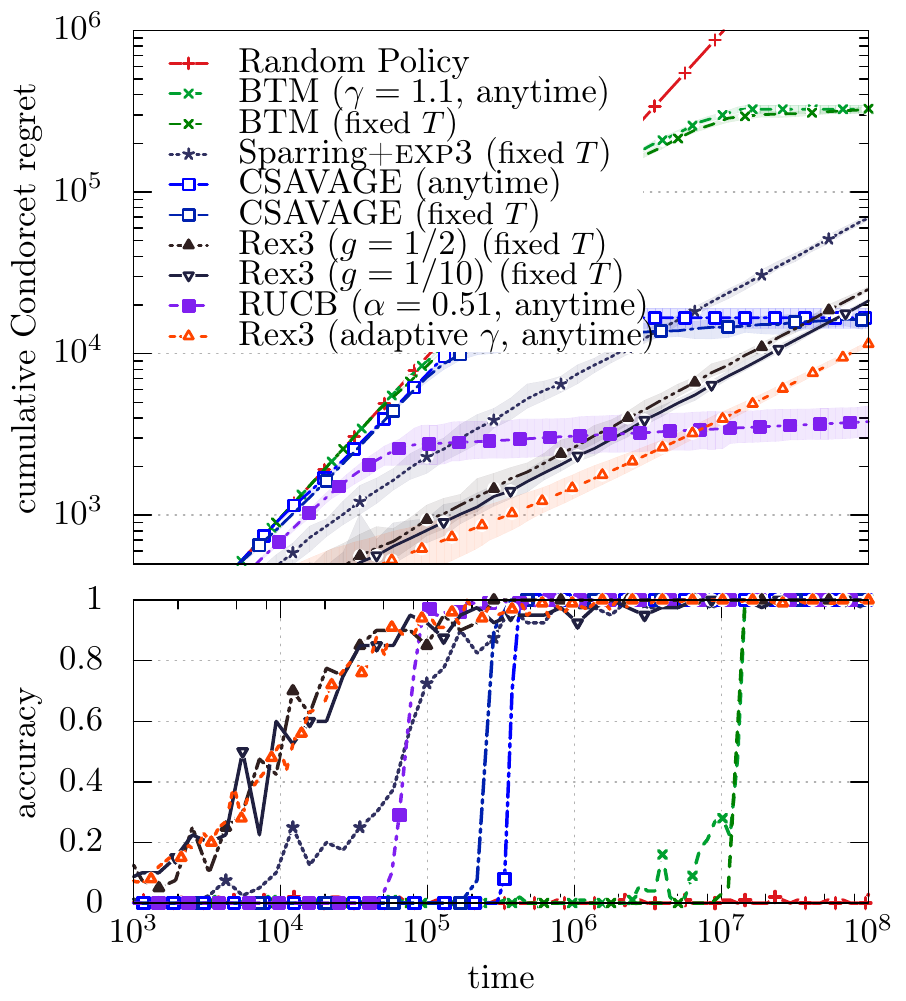}&
 \includegraphics[width=0.45\textwidth]{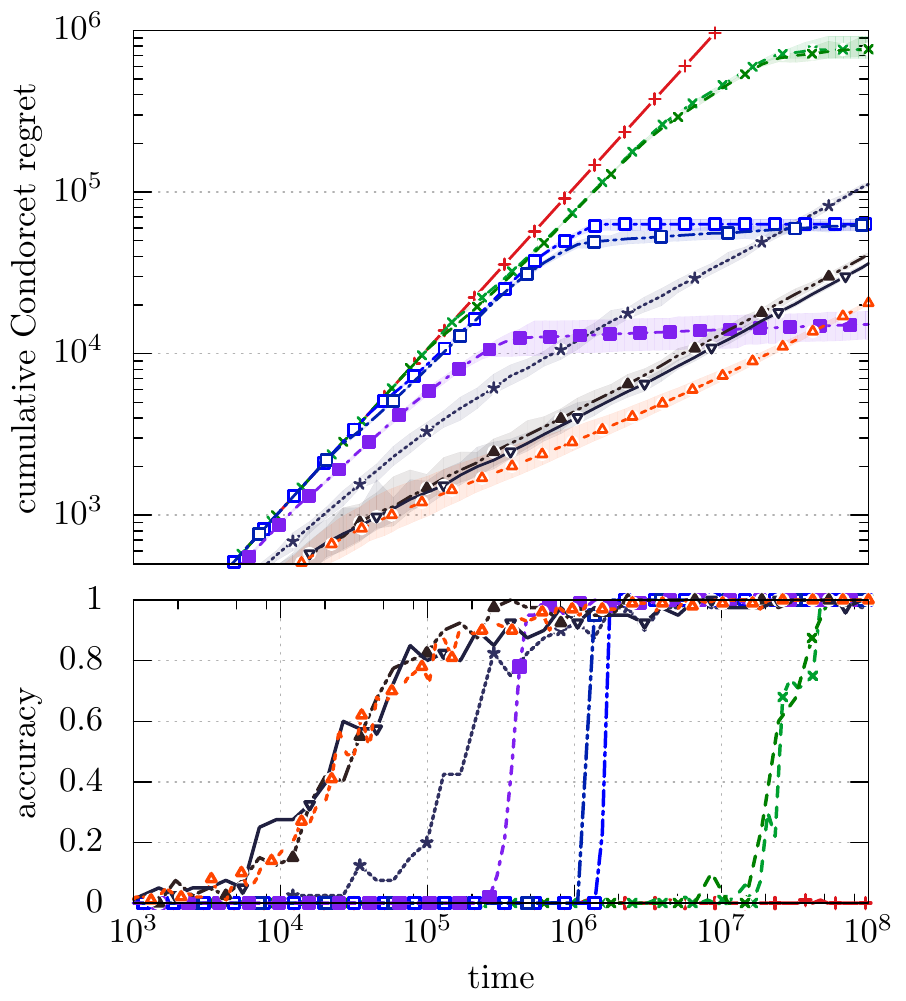}
 \end{tabular}
\caption{Average regret and accuracy plots on {\sc letor np2004} with respectively $16$, and $32$ rankers. We also give the anytime runs for {\sc btm} and {\sc savage} with a conservative $\delta=10^{-8}$.
{\it On regret plots, both time and regret scales are logarithmic ($\sqrt{t}$ hence appears as $t/2$). 
The colored areas around the curves show the minimal and maximal values over the runs}.
}
\label{fig:np2004bis}
\end{center}
\end{figure*}
\begin{figure*}
\begin{center}
\begin{tabular}{cc}
 \includegraphics[width=0.45\textwidth]{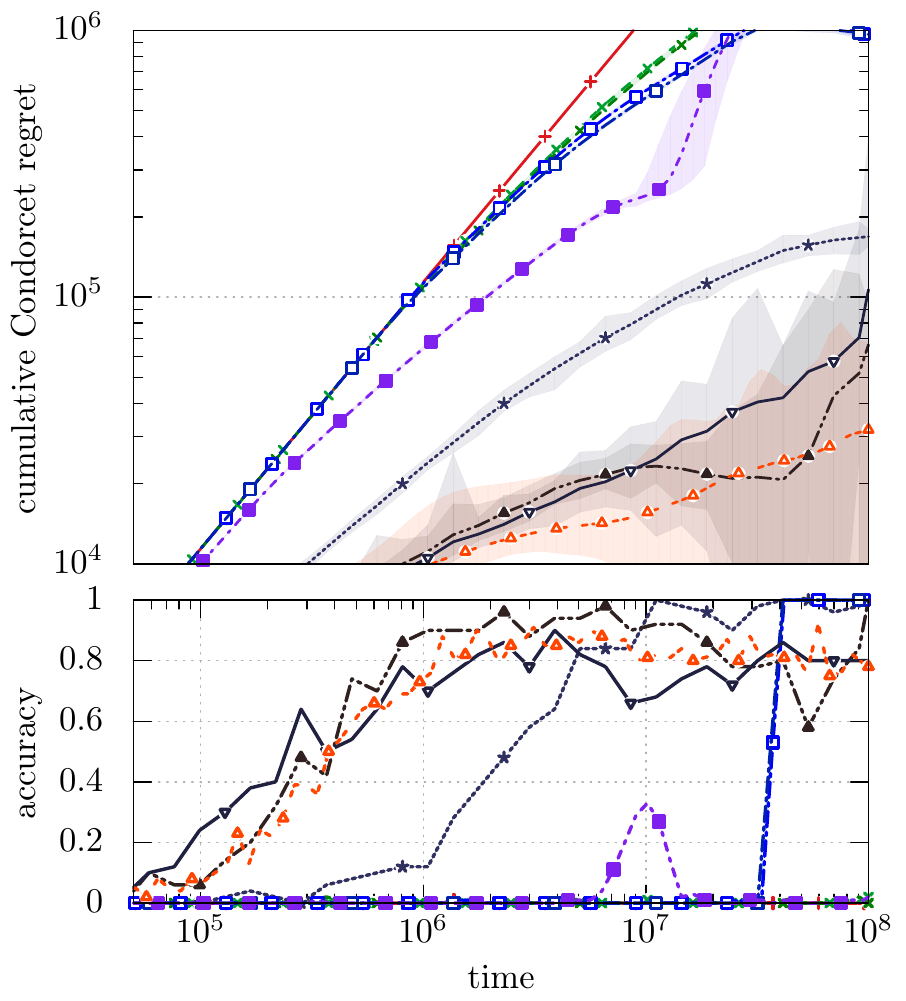}&
 \includegraphics[width=0.45\textwidth]{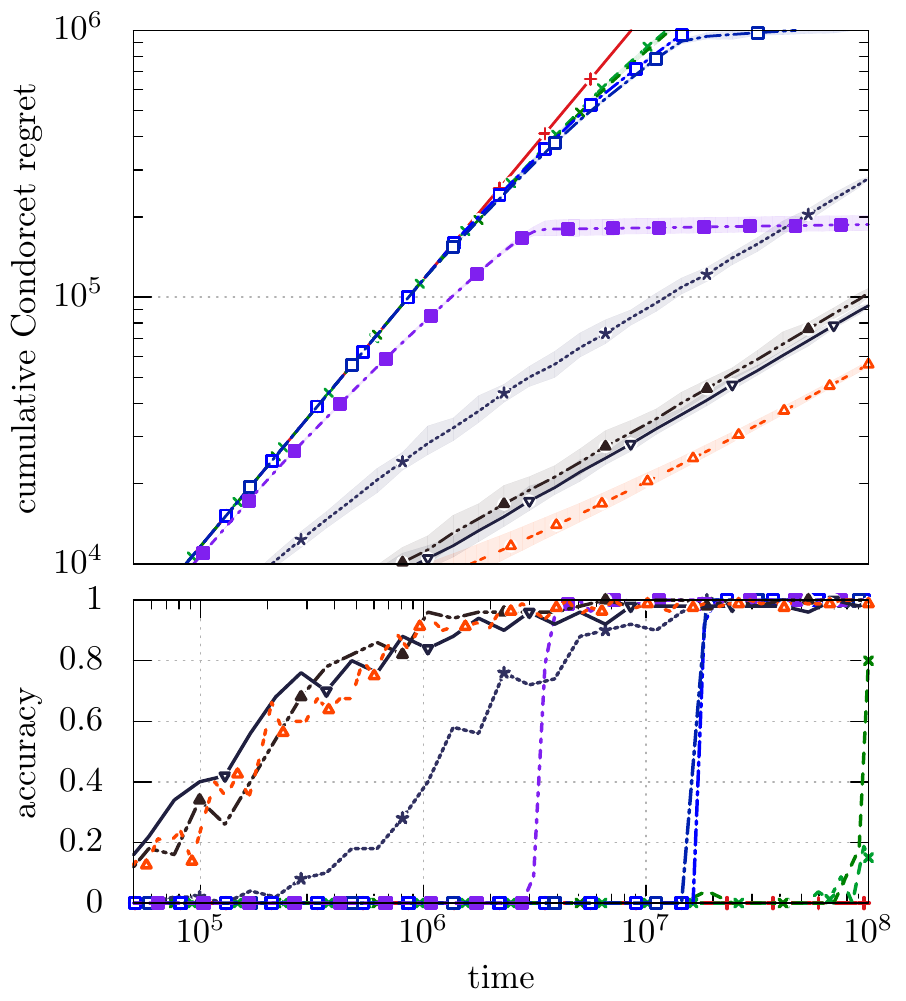}
 \end{tabular}
\caption{Expected regret and accuracy plots on {\sc mslr30k} with respectively informational and perfect navigational queries (136 rankers). There is no Condorcet winner on the left-hand-side informational queries matrix (we took a Copeland winner as a placeholder but the regret is negative for some arms).}
\label{fig:mslr30kbis}
\end{center}
\end{figure*}

\begin{figure*}
\begin{center}
\begin{tabular}{cc}
 \includegraphics[width=0.45\textwidth]{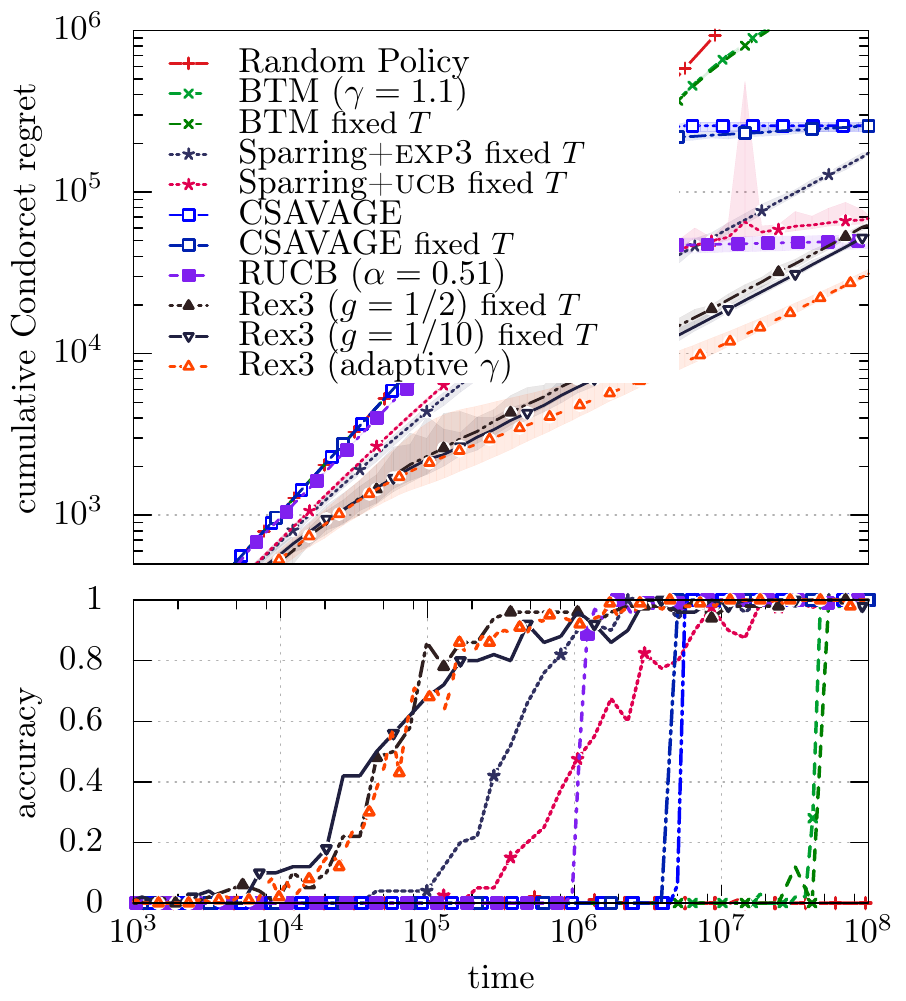}&
 \includegraphics[width=0.45\textwidth]{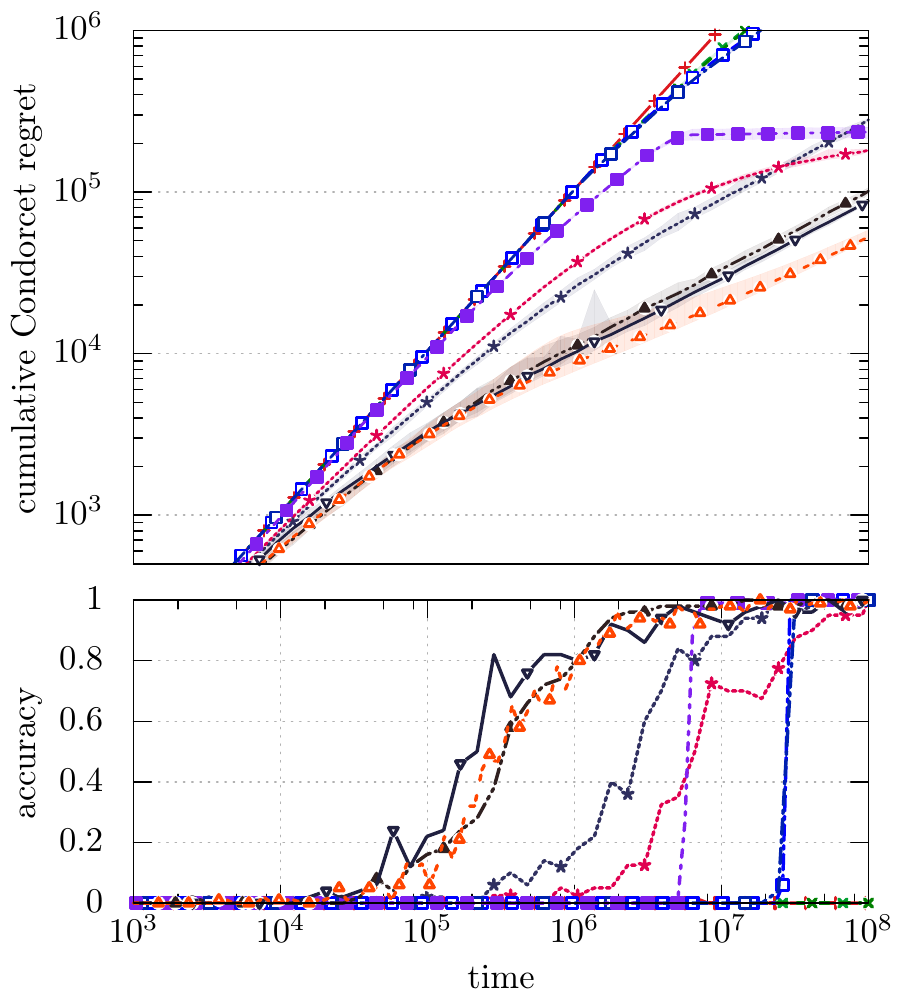}
 \end{tabular}
\caption{Average regret and accuracy plots respectively on {\sc letor np2004} (64 rankers) and
{\sc mslr30k} navigational queries (136 rankers) with Sparring coupled with a standard {\sc ucb} {\sc mab}.}
\label{fig:sparringUCB}
\end{center}
\end{figure*}

\begin{figure*}
\begin{center}
\begin{tabular}{cc}
 \includegraphics[width=0.45\textwidth]{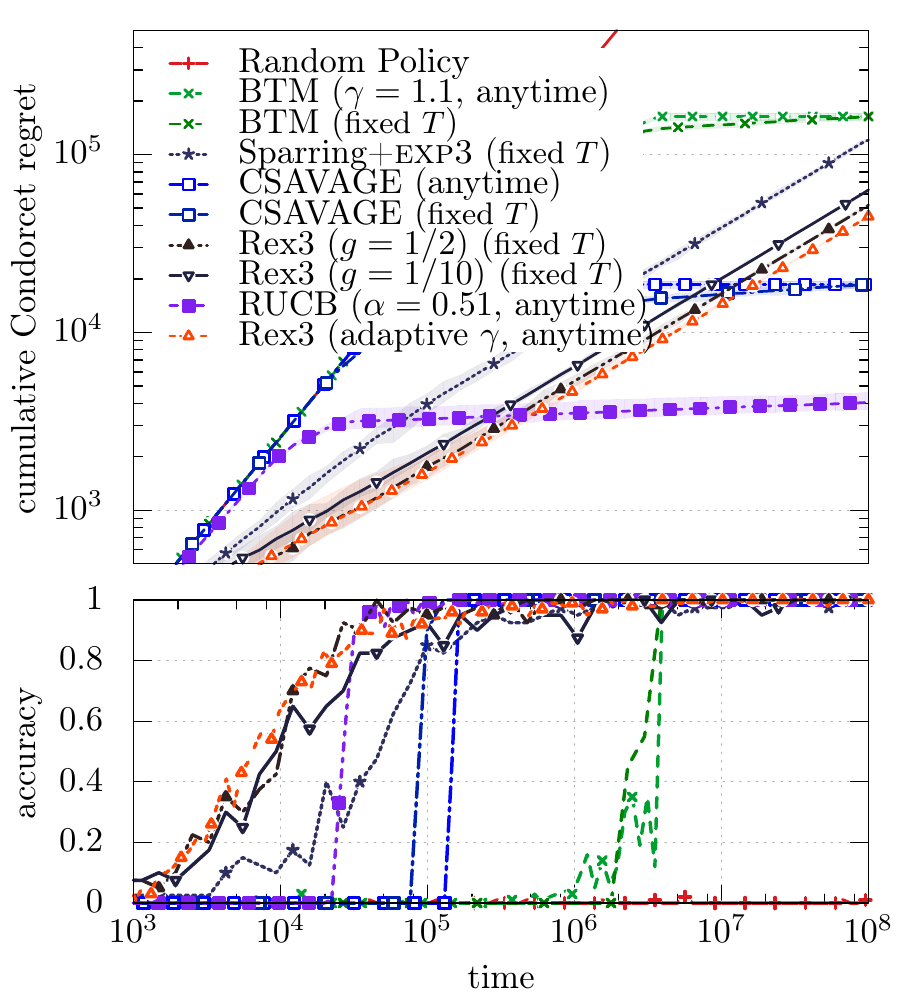}&
 \includegraphics[width=0.45\textwidth]{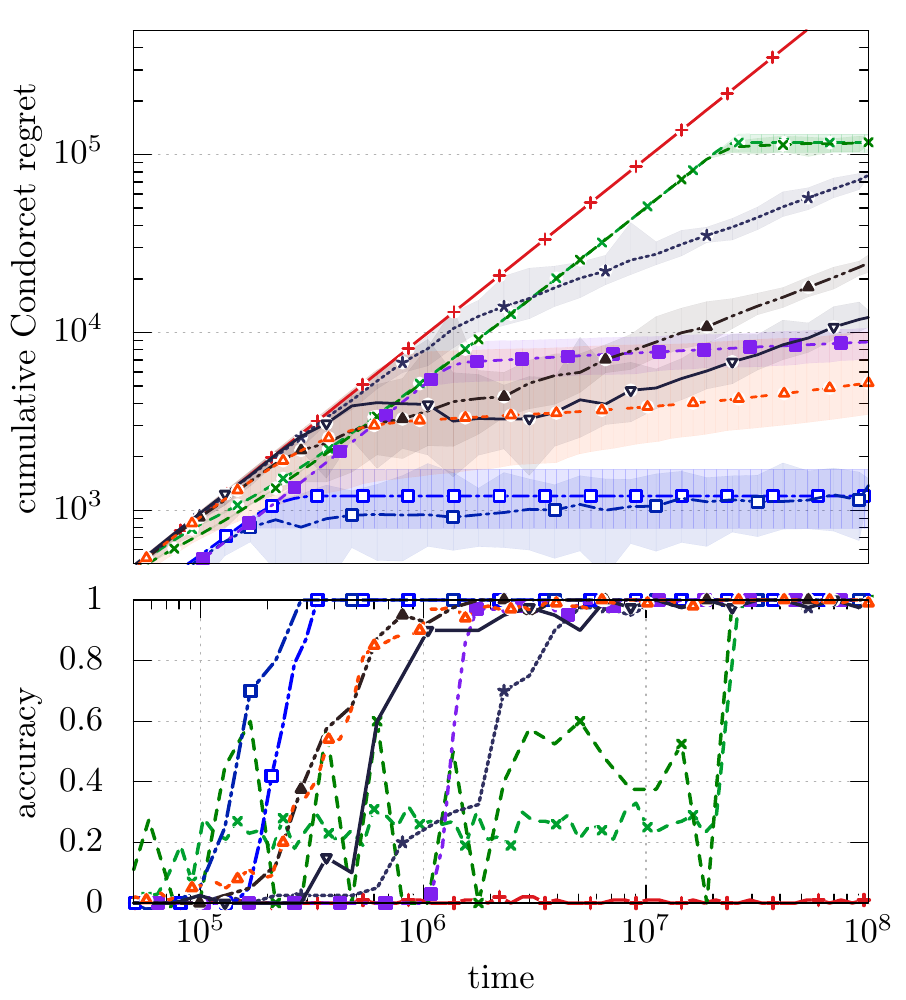}
 \end{tabular}
 \caption{Average regret and accuracy plots respectively on {\sc savage} and {\sc Bvs} artificial matrices.}
\label{fig:artificial}
\end{center}
\end{figure*}
\begin{figure*}
\begin{center}
\begin{tabular}{cc}
 \includegraphics[width=0.45\textwidth]{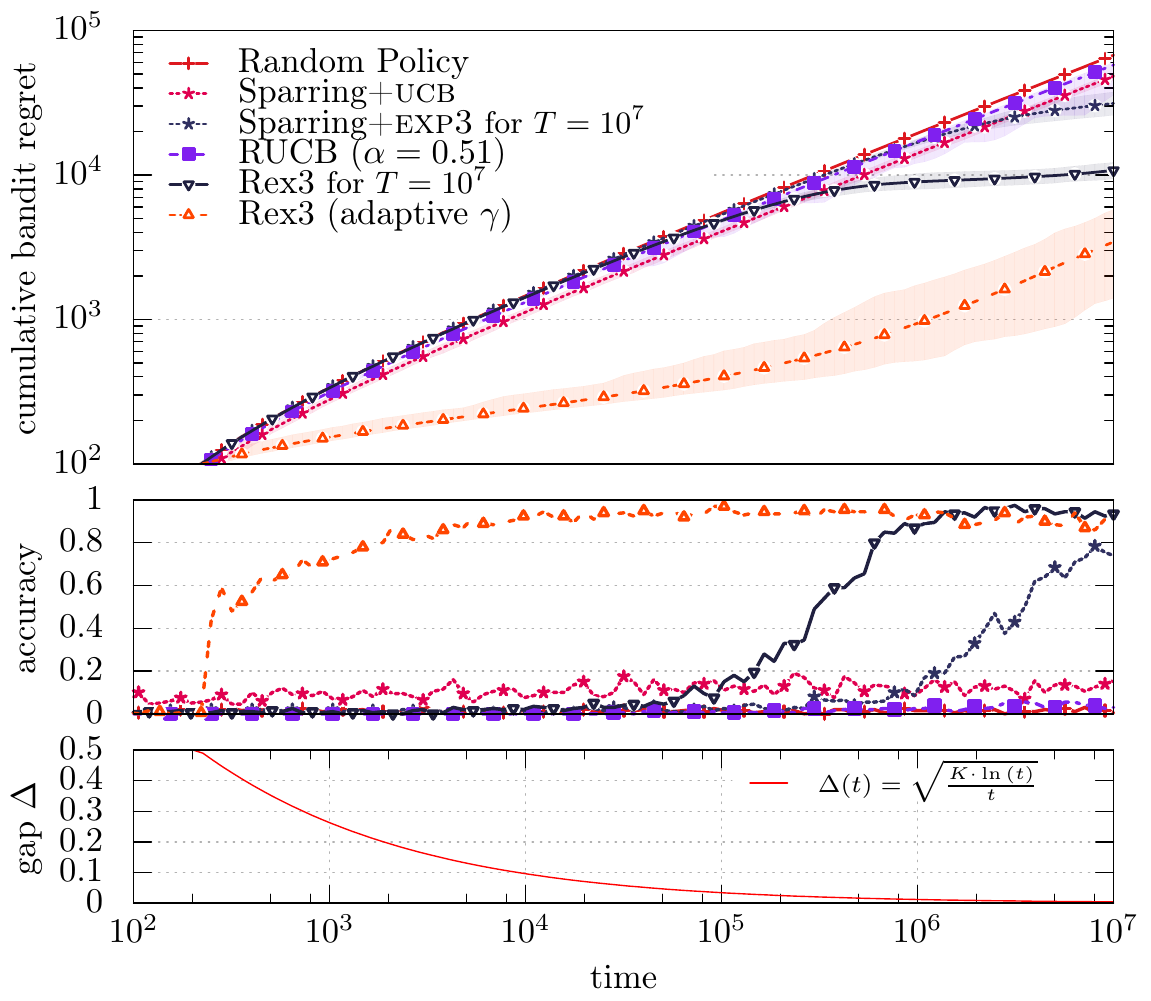}&
 \includegraphics[width=0.47\textwidth]{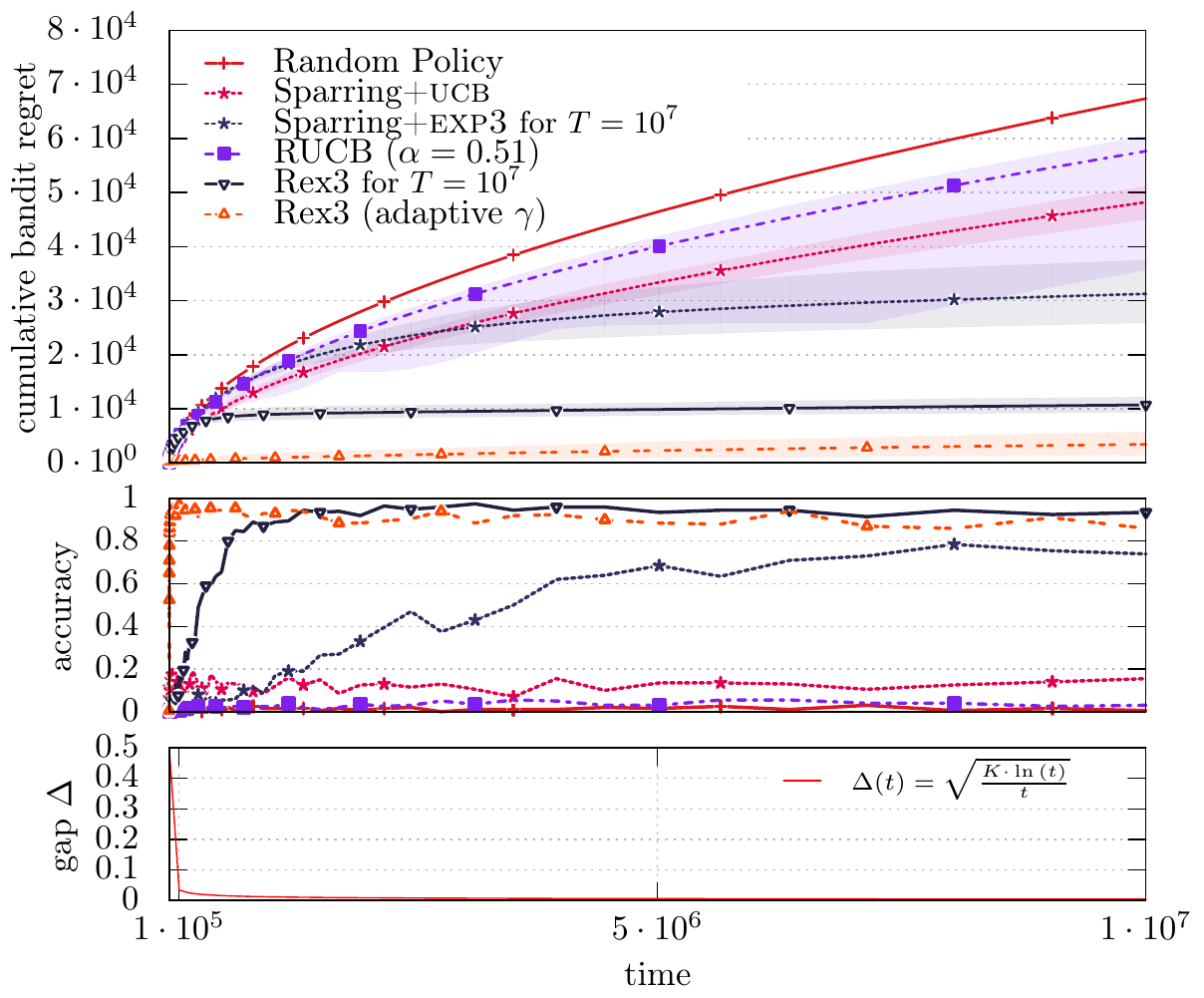}
 \end{tabular}
 \caption{In this plot, we experiment a synthetic utility-based $10$-armed dueling bandits problem with non-stationary rewards. The rewards are taken from Bernoulli distributions.
The best arm has a time-dependent expected reward equal to
 $1/2+\Delta(t)$ with $\Delta(t)=\sqrt{K\cdot{}\log(t)/t}$. The others arms' rewards are stationary with a mean of $1/2$. The gap function $\Delta(t)$ has been chosen to deceive stochastic algorithms: $\mathcal{O}\left(\frac{K\cdot\log(T)}{\Delta(T)}\right)\sim\mathcal{O}\left(\sqrt{KT\cdot\log(T)}\right)$.
{\it To ease reading we provide the same plot with logarithmic scale on the left and linear scale on the right.}}
 \label{fig:adv}
\end{center}
\end{figure*}
\begin{table}
\begin{center}
\caption{Some preference matrices we used.}
\label{tab:matrices}
\begin{tabular}{l@{\ }r@{\ }c@{\quad}c}
\bf Dataset & \bf $K$ & \bf Condorcet & \bf =Borda?\\ \Xhline{4\arrayrulewidth}
\sc ARXIV 2011 & 6 & yes & yes \\  \hline 
\sc LETOR NP2004 & 16 & yes & yes \\  \hline 
\sc LETOR NP2004 & 32 & yes & yes \\  \hline 
\sc LETOR NP2004 & 64 & yes & yes \\  \hline 
\sc MSLR Inf. & 136 & \bf no & - \\  \hline 
\sc MSLR Nav. & 136 & yes & yes \\  \hline 
\sc MSLR Perf. & 136 & yes & yes \\  \Xhline{4\arrayrulewidth}
%\sc Swirl {\small (artificial)}& 20 & yes & yes \\  \hline 
\sc SAVAGE {\small (artificial)}& 6 & yes & yes \\  \hline  
{\sc SAVAGE} {\small (artificial)}& 30 & yes & yes \\  \hline 
{\sc Bvs} {\small (artificial)}& 20 & yes & \bf no \\  \hline 
\end{tabular}
\end{center}
\end{table}

}{} % toggle long-version

\end{document}